\documentclass{article}

\usepackage{microtype}
\usepackage{graphicx}
\usepackage{booktabs} %

\usepackage[frozencache,cachedir=.]{minted}
\usepackage[scaled=0.825]{beramono}
\usepackage[table,x11names,dvipsnames]{xcolor}
\usepackage{hyperref}

\usepackage[accepted]{icml2025}

\usepackage{amsmath}
\usepackage{amssymb}
\usepackage{mathtools}
\usepackage{amsthm}

\usepackage[capitalize,noabbrev]{cleveref}

\theoremstyle{plain}

\theoremstyle{definition}

\theoremstyle{remark}

\usepackage[textsize=tiny]{todonotes}
\usepackage{wrapfig}
\usepackage{multirow}
\definecolor{BlueViolet}{rgb}{0.54, 0.17, 0.89}

\newcommand{\code}[2][BlueViolet]{\textcolor{#1}{\texttt{#2}}}
\newcommand{\lora}{LoRA}
\newcommand{\ours}{\code[BlueViolet]{T2L}}
\usepackage{tcolorbox}
\DeclareMathOperator*{\argmin}{argmin}
\newcommand{\Larch}{\colorbox{SteelBlue1!80}{\textbf{L}}}
\newcommand{\March}{\colorbox{MediumPurple1!60}{\textbf{M}}}
\newcommand{\Sarch}{\colorbox{Pink1}{\textbf{S}}}
\newcommand{\uptriangle}{{\color{ForestGreen}$\blacktriangle$}}
\newcommand{\downtriangle}{{\color{Red}$\blacktriangledown$}}
\newcommand{\greenbox}[1]{\colorbox{LimeGreen}{\textbf{#1}}}
\usepackage{subfig}

\icmltitlerunning{Text-to-LoRA: Instant Transformer Adaption}

\begin{document}

\twocolumn[
\icmltitle{Text-to-LoRA: Instant Transformer Adaption}

\icmlsetsymbol{equal}{*}

\begin{icmlauthorlist}
\icmlauthor{Rujikorn Charakorn}{sakana}
\icmlauthor{Edoardo Cetin}{sakana}
\icmlauthor{Yujin Tang}{sakana}
\icmlauthor{Robert T. Lange}{sakana}
\end{icmlauthorlist}

\icmlaffiliation{sakana}{Sakana AI}

\icmlcorrespondingauthor{Rujikorn Charakorn}{rujikorn@sakana.ai}
\icmlcorrespondingauthor{Robert T. Lange}{robert@sakana.ai}

\icmlkeywords{Machine Learning, ICML}

\vskip 0.3in
]

\printAffiliationsAndNotice{}  %

\begin{abstract}
While Foundation Models provide a general tool for rapid content creation, they regularly require task-specific adaptation. Traditionally, this exercise involves careful curation of datasets and repeated fine-tuning of the underlying model. 
Fine-tuning techniques enable practitioners to adapt foundation models for many new applications but require expensive and lengthy training while being notably sensitive to hyperparameter choices.
To overcome these limitations, we introduce Text-to-LoRA (\ours{}), a model capable of adapting large language models (LLMs) \emph{on the fly} solely based on a natural language description of the target task.  \ours{} is a hypernetwork trained to construct \lora{}s in a single inexpensive forward pass. 
After training \ours{} on a suite of 9 pre-trained \lora{} adapters (GSM8K, Arc, etc.), we show that the ad-hoc reconstructed \lora{} instances match the performance of task-specific adapters across the corresponding test sets.
Furthermore, \ours{} can compress hundreds of \lora{} instances and zero-shot generalize to entirely unseen tasks.
This approach provides a significant step towards democratizing the specialization of foundation models and enables language-based adaptation with minimal compute requirements.
Our code is available at \url{https://github.com/SakanaAI/text-to-lora}.
\end{abstract}

\section{Introduction}

\begin{figure*}[th]
    \centering
    \hspace*{\fill}
    \begin{minipage}{0.45\linewidth}
        \centering
        \includegraphics[trim={0cm 0cm 18.5cm 0cm}, width=0.95\linewidth, clip]{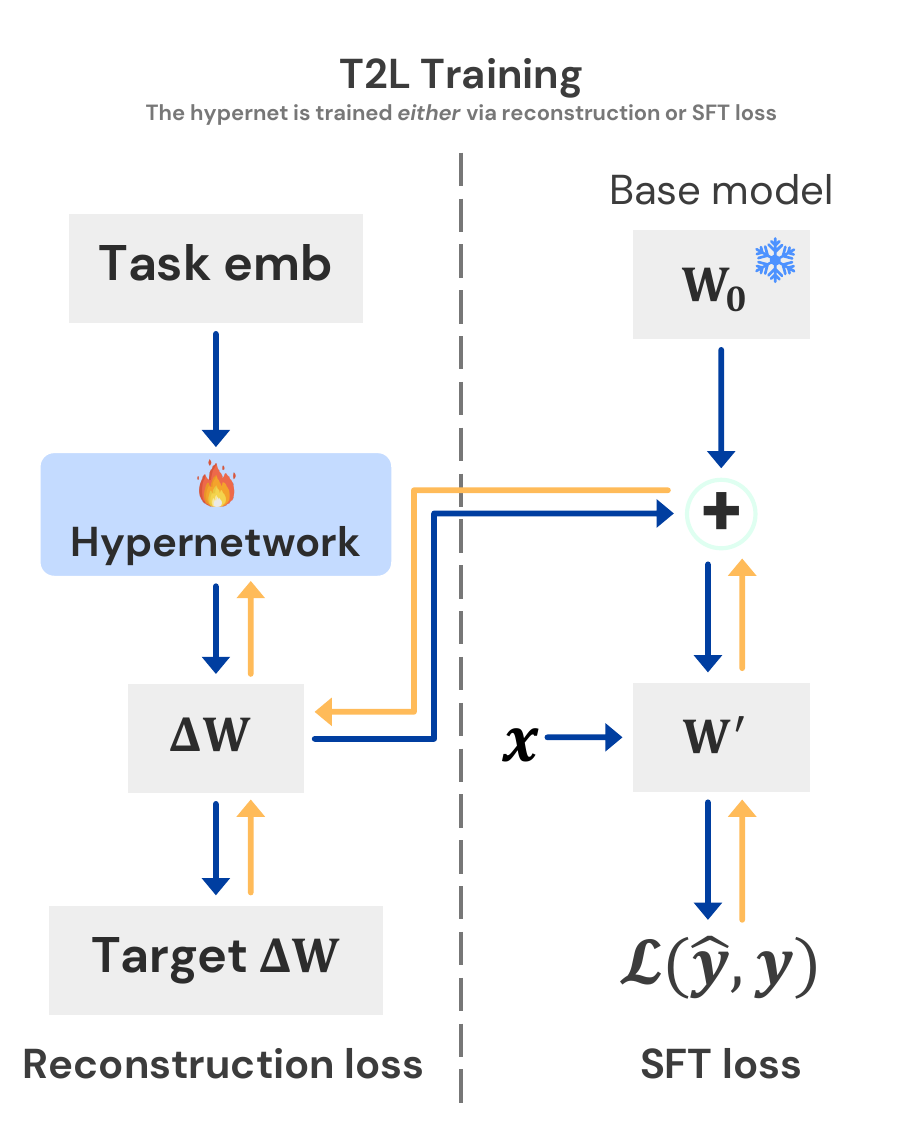}
        \label{subfig:hyperlora}
    \end{minipage}
    \begin{minipage}[c]{0.35\linewidth}
        \centering
        \includegraphics[width=0.95\linewidth]{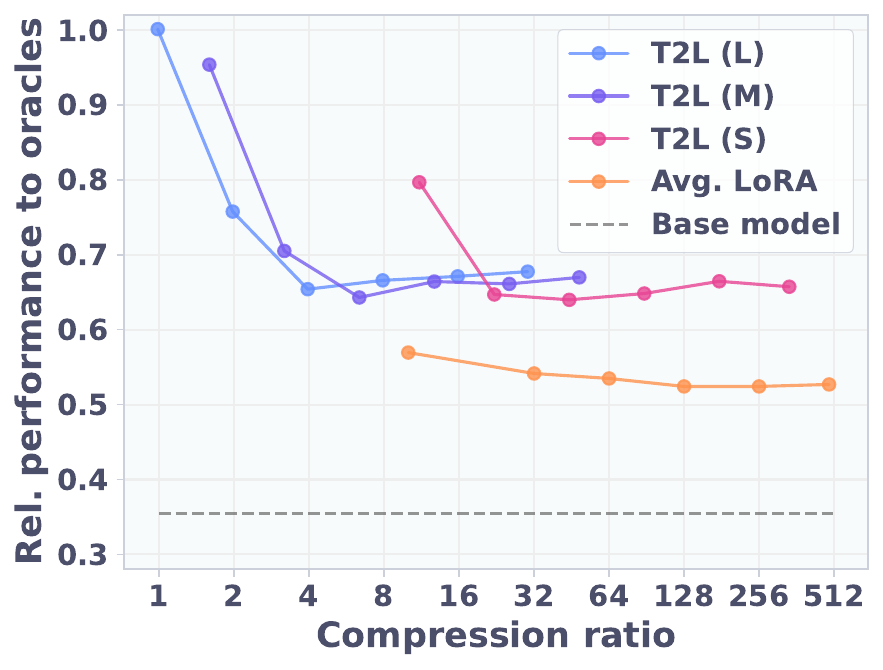}
        \label{subfig:compression_scale}
        \\
        \includegraphics[width=0.95\linewidth]{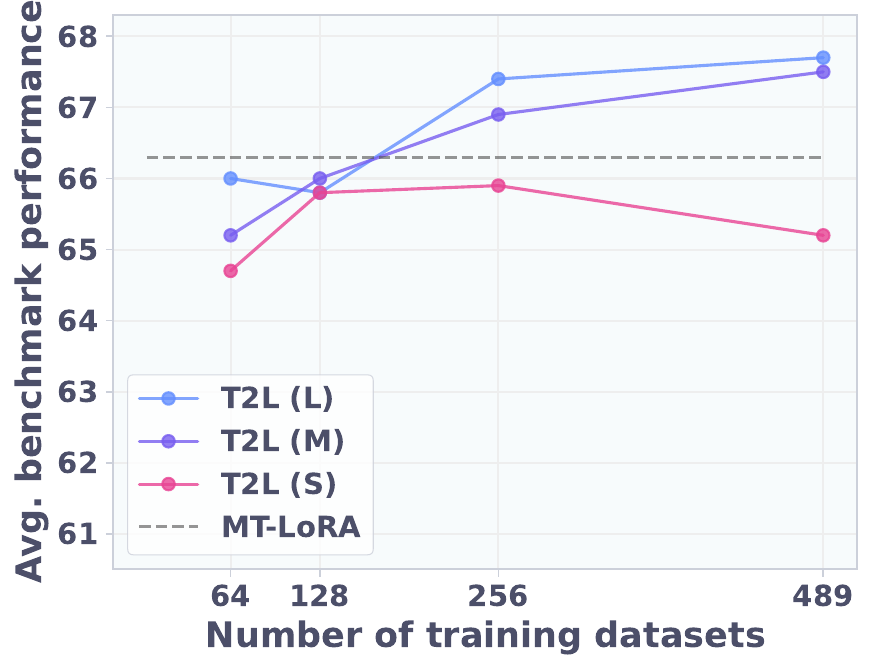}
        \label{subfig:main_fig_scaling_training_tasks}
    \end{minipage}
    \hspace*{\fill}
    \caption{\textbf{Left:} Conceptual overview of \ours{}'s training routine. Given a set of task description embeddings, we train a hypernetwork to generate \lora{} adaptation matrices ($\Delta W$) for various tasks. The weights of \ours{} are \emph{either} optimized to distill pre-trained \lora{} weights or via multi-task supervised fine-tuning on downstream tasks. \textbf{Right, Top:} Relative performance to the oracles on training SNI tasks with varying compression ratios. \textbf{Right, Bottom:} Zero-shot \lora{} generation performance on 10 benchmark tasks. As we increase the number of pre-training datasets, the performance of \ours{} increases for 3 different \ours{} architectures.}
    \label{fig:conceptual}
\end{figure*}

Biological systems are capable of rapid adaptation, given limited sensory cues. For example, the human visual system can tune its light sensitivity and focus through neuromodulation of the fovea and rod cells \citep{wurtz2011neuronal, digre2012shedding}. 
While recent LLMs exhibit a wide variety of capabilities and knowledge, they remain rigid when adding task-specific capabilities. 
In such cases, practitioners often resort to re-training parts of the model \citep{gururangan2020don, wei2021finetuned, dettmers2022gpt3, tay2021scale} using parameter-efficient fine-tuning techniques, e.g., Low-Rank Adaptation \citep[\lora{},][]{hu2022lora}.
Typically, a \lora{} adapter has to be optimized for each downstream task and requires task-specific dataset and hyperparameter setting. This fine-tuning scheme for adaptation significantly limits the possibility of transferring knowledge between tasks and induces engineering overhead.
Recently, it has been observed that by inducing structural constraints, the low-rank matrices learned by \lora{} adapters can be further compressed. For example, one can train \emph{lossy} versions of the original adapter while maintaining downstream performance \citep{bruel2024compress_then_serve, kim2024cond_lora, kopiczko2024vera}. Furthermore, multiple \lora{}s can be combined for new tasks at inference time \citep{ostapenko2024towards_modular_llms}. 
At the core of these approaches lies the explicit use of decomposition or dimensionality reduction techniques (e.g., SVD or routing) for better compression and online composition of existing \lora{}s. This raises the following questions: 

\begin{tcolorbox}[colback=blue!5, colframe=blue!50, boxrule=1mm]
\begin{enumerate}
    \item Can we end-to-end train a neural network to compress many pre-trained \lora{}s? 
    \item Can we decode new task-specific \lora{} adapters solely based on natural-language instructions for an unseen task at test time?
\end{enumerate}
\end{tcolorbox}

We hypothesize that different \lora{} adapters share the same underlying adaptation mechanism and can be optimized simultaneously without any explicit structure or recipe for combining them. To explicitly test this hypothesis, we propose \ours{}, a hypernetwork \citep{ha2016hypernetworks} that compresses task-specific \lora{}s and generates new \lora{} adapters \emph{zero-shot} at inference time. \ours{} is trained to compress \lora{}s on a diverse task distribution from the Super Natural Instructions (SNI) dataset \citep{wang2022sni}.
Importantly, \ours{} takes a natural language description of the target task as an input, allowing zero-shot \lora{} generation to unseen tasks.
Empirically, we show that \ours{} can effectively be trained either to reconstruct pre-trained adapters or via supervised fine-tuning on a distribution of downstream tasks (see \cref{fig:conceptual}, top right).
After training, \ours{} outperforms a multi-task \lora{} baseline and Arrow Routing~\citep{ostapenko2024towards_modular_llms}, a state-of-the-art zero-shot \lora{} routing method, on various benchmark tasks.
Furthermore, we show that \ours{} can generate \lora{} adapters for previously unseen tasks solely using the language-based task description. 
This result highlights the generalization capabilities and applicability of our proposed indirect adaptation encoding. Our contributions are summarized as follows:
\begin{enumerate}
    \item We introduce hypernetwork-based architectures for producing \lora{} adapters with a single forward pass (\cref{sec:architecture}) based on text descriptions. \ours{} architectures can be trained using both distillation of pre-trained adapters and supervised multi-task fine-tuning.
    \item We show that \ours{} can efficiently encode hundreds of \lora{} adapters (\cref{sec:experiments}). While the compression is lossy, \ours{} maintains the performance of task-specifically tuned \lora{} adapters. Furthermore, \ours{} can generalize to unseen tasks given suitable natural language descriptions of the tasks.
    \item We provide rigorous ablations (\cref{sec:ablations}) including \ours{} scaling with datasets (see \cref{fig:conceptual}, bottom right), the impact of different task description embeddings, the training routines, and text-based task descriptions.
    \item Finally, we analyze the nature of \ours{} generations.
    We find semantically meaningful \lora{} clusters when visualizing the generated \lora{}s in a dimensionality-reduced space (\cref{subsec:umap_visualization}).
    Furthermore, we study the relationship between \lora{} adapters and find compelling evidence why reconstruction-trained \ours{} cannot generalize (\cref{subsec:lora_of_similar_tasks}). 
\end{enumerate}

\section{Preliminaries}
\begin{figure*}[ht]
    \centering
    \includegraphics[trim={0cm 7.7cm 3.1cm 0cm},clip,width=0.6\linewidth]{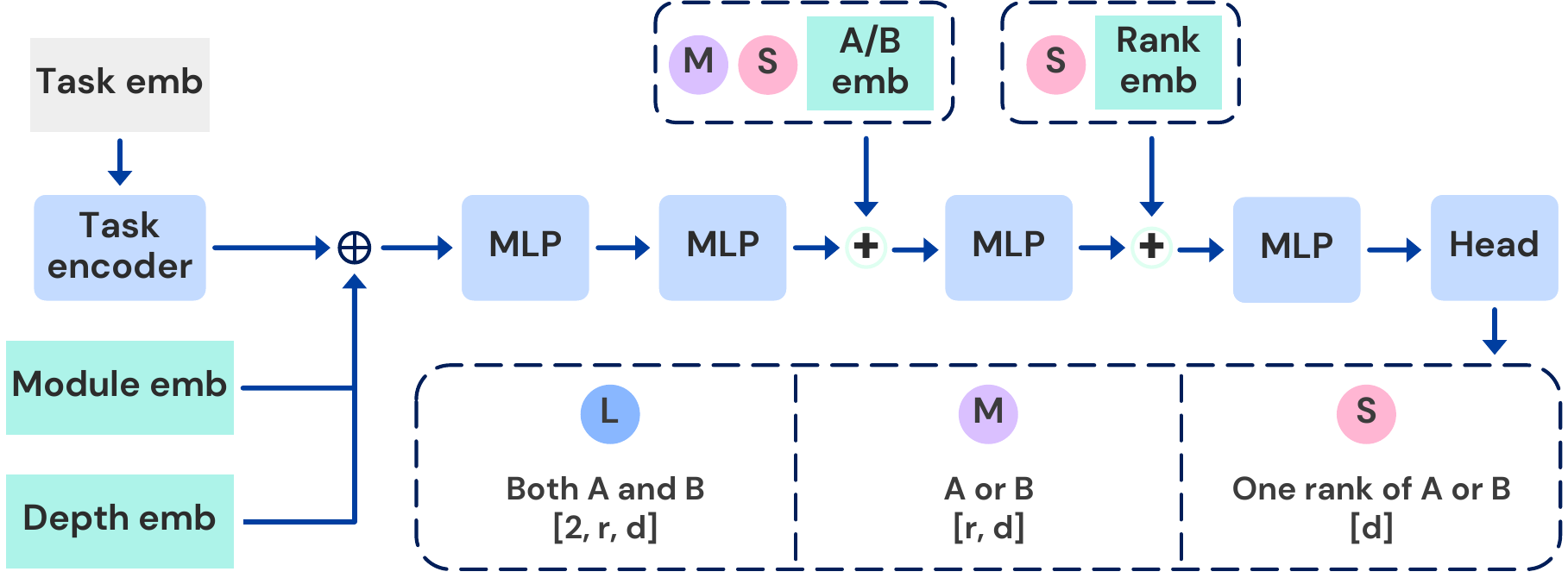}
    \caption{Overview of \ours{} architectural variations. The dashed box at the bottom shows the output size of a single forward pass of \ours{}. Blue boxes are trainable modules. Cyan boxes are trainable embedding layers. Components in dashed boxes are only used with their corresponding architectures. $r$ is the rank of a \lora{} adapter and $d$ is the size of the input and the output dimension.}
    \label{fig:architectures}
    \vspace{-0.2cm}
\end{figure*}
We utilize multiple fine-tuning datasets $\mathcal{D} = \{\mathcal{D}^1, \dots, \mathcal{D}^T\}$, which correspond to different tasks $\mathcal{T} = \{t^1, \dots, t^T\}$. For the purpose of training \ours, we assume that each fine-tuning dataset has a set of natural language \emph{task descriptions} ($Z^i = \{z^i_1, \dots, z^i_m\}$): $\mathcal{D}^i = \{ X^i, Y^i, Z^i\}$. The task descriptions do not need to be specific to each sample but rather a general description of the dataset. For a single task $t^i$, the fine-tuning objective of an LLM with pre-trained weights ($\Psi$) is given by
\begin{align}
    \Delta W^i = \argmin_{\Delta W^i} \mathcal{L}_\text{SFT}(\mathcal{D}^i, \Psi, \Delta W^i),
\end{align}
where $\mathcal{L}_\text{SFT}$ gives the supervised fine-tuning loss and $\Delta W^i$ is the fine-tuning adaption for task $t^i$ to the base weights.
For the \emph{multi-task} setting, we train a single adapter $\Delta W$ to minimize the expected loss over the union of all datasets $\mathcal{D}$:
\begin{align}
    \Delta W = \argmin_{\Delta W} \mathbb{E}_{\mathcal{D}^i \sim \mathcal{D}}  \; \mathcal{L}_\text{SFT}(\mathcal{D}^i, \Psi, \Delta W).
\end{align}
\textbf{Low-Rank Adaptation \citep[\lora{}, ][]{hu2022lora}:}
\lora{} is a parameter-efficient fine-tuning method that freezes the pre-trained weights of a base model and only learns low-rank weight matrices, which serve as an adapter to the base model. For each selected linear transformation $h = W_0 x$, the fine-tuned transformation is given by
$h = W_0 x + \Delta W x = W_0 x + B^T A x$,
where $A, B \in \mathbb{R}^{r \times d}$ are weight matrices of rank $r < d$. We omit the layer index and module type of the \lora{} weights when referring to all \lora{} weights. Otherwise, we use subscripts to represent the layer index and module type, e.g., $\Delta W_{m,l}$, where $m$ is the module type (e.g., query projection) and $l$ is the layer index.

\textbf{Hypernetworks:} 
A hypernetwork is a neural network that generates parameters for another `base' network \citep{ha2016hypernetworks}. It serves as an indirect encoding \citep{schmidhuber1997discovering, stanley2003taxonomy,zhang2018graph,schug2024attention} of the base network, given that the parameter count of the hypernetwork is much smaller. This compression is achieved by learning to share parameters indirectly. More specifically, given a layer-specific descriptor vector $\phi_l$, a hypernetwork with parameters $\theta$ generates the parameters of the base model at layer $l \in \{1, \dots L\}$ as follows: $W_l = h_\theta(\phi_l)$.
Traditionally, the layer descriptors are either one-hot or learned vectors.
The weights $\theta$ are then trained via end-to-end optimization on a downstream task.

\section{Text-to-LoRA: Learning to Compress and Generate LoRAs}
\label{sec:architecture}
In this work, we utilize a hypernetwork to generate \lora{} adapters for task-specific adaptation. For each target module ($m$) and layer index ($l$), a hypernetwork generates the two low-rank matrices $A, B$ based on a task description $z^i \in Z^i$ of a task $t^i$ as follows:
\begin{align}
&\Delta W^i_{m,l} = h_\theta(\phi^i_{m,l}), \text{  with  } \\ &\phi^i_{m,l}  = \texttt{concat}\left[f(z^i), E[m], E[l]]\right],
\end{align}
where $f$ gives a vector representation of a text description, typically represented by a \texttt{CLS} token of a bidirectional transformer model or last token activation of an LLM. $E$ is a learnable embedding dictionary indexed by either a module type $m$ or a layer index $l$.
For legibility, we introduce a shorthand notation for \ours{}'s output
$\Delta W^i \coloneq h_\theta(\phi^i) \coloneq h_\theta(\{\phi^i_{m,l}\})$.
Then, a supervised fine-tuning training objective for \ours{} is
\begin{align}
    \theta &= \argmin_\theta \; \mathbb{E}_{\mathcal{D}^i \sim \mathcal{D}, z^i \sim Z^i} \; \mathcal{L}_\text{SFT}(\mathcal{D}^i, \Psi,  h_\theta(\phi^i)), \label{eq:sft_hyperlora}
\end{align}
Note that values of $m$ and $l$ can be batched, which allows \ours{} to generate $\Delta W$ for all modules and layer indices efficiently within a single forward pass.

\subsection{Text-to-LoRA Architectures}
Most of a hypernetwork's parameters come from the output layer, which scales linearly with the size of the target weights \citep{von2019continual}.
To explore the complexity-performance trade-off, we propose three variants of \ours{}: \Larch, \March, and \Sarch.
We impose different output spaces on the hypernetwork that represent different inductive biases and parameter counts (see \cref{fig:architectures}). We note that all variants use the same backbone architecture and only differ in their output heads and learnable embeddings.
The \textbf{\Larch{} architecture} is the largest variant. Its final linear layer outputs low-rank $A$ and $B$ matrices simultaneously, with the number of weight connections to the output head $|\theta_\text{head}| = {d_\text{out} \times 2 \times r \times d}$, where $d_\text{out}$ is the output size of the last MLP block.
\textbf{\March{} architecture} is the medium-sized model with a shared output layer between the low-rank $A$ and $B$ matrices. That is, the head outputs a low-rank matrix, either $A$ or $B$, depending on the learnable embedding. The size of the output head is $|\theta_\text{head}| = {d_\text{out} \times r \times d}$.
Finally, \textbf{\Sarch{} architecture} is the most parameter-efficient model with the strongest inductive biases, where the hypernetwork outputs only one rank of a low-rank matrix at a time. This output space makes the size of the head much smaller: $|\theta_\text{head}| = {d_{emb} \times d}$.
For reference, a \lora{} adapter has $r \times d \times 2 \times L \times |M|$ trainable parameters, where $L$ is the number of layers and $|M|$ is the number of target modules.
The default value of $d_\text{out}$ is $512$.
We note that every architecture can generate all the low-rank matrices $A$ and $B$ in a single forward pass by batching all the input embeddings.
We provide more details of the architectures in \cref{app:architecture} and the weight initialization method that leads to stable training in \cref{app:hyper_init}.

\subsection{Training Text-to-LoRA via LoRA Reconstruction}
The most straightforward way to train \ours{} is to reconstruct pre-trained task-specific \lora{}s. This setup allows us to utilize publicly available libraries of \lora{}s \citep{bruel2024compress_then_serve, zhao2024loraland}. Alternatively, one can also use a two-stage procedure, in which a library of \lora{}s is pre-trained in the first stage and then train \ours{} to reconstruct them.
For the sole purpose of compressing \lora{}s, we can train \ours{} using one-hot or learnable vectors as task embeddings. 
However, these embeddings do not allow zero-shot \lora{} generation for unseen tasks. To enable zero-shot \lora{} generation, we additionally condition \ours{} with embeddings of natural language task descriptions, which allows \ours{} to generate \lora{} adapters for various tasks---including unseen ones---given corresponding task descriptions. Given a suitable library of \lora{} adapters $\Omega$, the reconstruction loss for \ours{} can be written as
\begin{align}
    \mathcal{L}(\Omega, \theta) = \mathbb{E}_{\Delta W^i \sim \Omega} \; |\Delta W^i - h_\theta(\phi^i)|.
\end{align}
\vspace{-0.5cm}
\subsection{Training Text-to-LoRA via Supervised Fine-Tuning}
Alternatively, \ours{} can be directly optimized on fine-tuning datasets. 
Training \ours{} with SFT sidesteps the need for intermediate target \lora{} adapters and allows for end-to-end training.
This training scheme is preferred if existing trained \lora{}s are not naturally clustered by their functionalities or downstream tasks. For instance, $t^1$ and $t^2$ could be two related tasks requiring a similar LLM capability, but $\Delta W^1$ and $\Delta W^2$ could be in different minima. Thus, \ours{} trained via reconstruction training would have to compress numerically different $\Delta W^1$ and $\Delta W^2$, making it less likely to generalize. In fact, we empirically find that a \ours{} trained via reconstruction fails to generalize to unseen tasks (\cref{subsec:recon_vs_sft_zeroshot}).
In contrast, an SFT-trained \ours{} can implicitly learn to cluster tasks, which has been shown to improve zero-shot \lora{} routing performance \citep{ostapenko2024towards_modular_llms}.
The SFT loss for \ours{} is given by \cref{eq:sft_hyperlora}.

\section{Experiments}
\label{sec:experiments}
We investigate the effectiveness of the different \ours{} architectures and training schemes in terms of the compression of adapters (\cref{sec:lora_distill}) and zero-shot \lora{} generation for unseen tasks (\cref{sec:zeroshot}). As baselines, we consider task-specific \lora{}s, element-wise averaged \lora{}, and multi-task \lora{}---a \lora{} adapter trained on all training tasks.
We also implement Hyperdecoders \citep{ivison2022hyperdecoders}---a hypernetwork that generates LoRAs on a per-sequence basis---based on our proposed architectures.
To boost the performance of the base models without fine-tuning, we utilize few-shot in-context learning \citep[ICL,][]{icl,dong2024icl_survey} and task description prepending, i.e., providing task description at the beginning of each query.
Additionally, we include results of Arrow Routing zero-shot performance from \citet{ostapenko2024towards_modular_llms}. Note that the performance can only be compared indirectly as it uses a different set of \lora{} adapters and training tasks. Furthermore, there are likely differences in the benchmark evaluation prompts.

In most experiments, we use \texttt{Mistral-7B-Instruct} \citep{jiang2023mistral} as the base LLM model except in \cref{tab:zero-shot-llama,tab:zero-shot-gemma} where \texttt{Llama-3.1-8B-Instruct} and \texttt{Gemma-2-2b-Instruct} are used as the base models, respectively. We use \texttt{gte-large-en-v1.5} \citep{li2023towards, zhang2024mgte} for extracting the task embedding from a natural language task description. All \lora{} adapters are of rank 8 and only target the query and the value projection modules in every attention block of the base LLM (totaling $3.4$M parameters). With this \lora{} configuration, \Larch{}, \March{}, and \Sarch{} have $55$M, $34$M, and $5$M trainable parameters respectively.

\begin{table*}[ht]
\centering
\caption{Benchmark performance of \ours{} trained via reconstruction loss on 9 benchmark tasks. \greenbox{Green highlight} indicates that \ours{} outperforms the benchmark-specific \lora{} adapters.}
\label{tab:benchmark-recon}
\resizebox{0.8\linewidth}{!}{%
\centering
\begin{tabular}{llccccccccc|c}
\toprule
\multicolumn{2}{l}{} &
  \textbf{\begin{tabular}[c]{@{}c@{}}ArcC\\ (acc)\end{tabular}} &
  \textbf{\begin{tabular}[c]{@{}c@{}}ArcE\\ (acc)\end{tabular}} &
  \textbf{\begin{tabular}[c]{@{}c@{}}BQ\\ (acc)\end{tabular}} &
  \textbf{\begin{tabular}[c]{@{}c@{}}GSM8K\\ (acc)\end{tabular}} &
  \textbf{\begin{tabular}[c]{@{}c@{}}HS\\ (acc)\end{tabular}} &
  \textbf{\begin{tabular}[c]{@{}c@{}}OQA\\ (acc)\end{tabular}} &
  \textbf{\begin{tabular}[c]{@{}c@{}}PIQA\\ (acc)\end{tabular}} &
  \textbf{\begin{tabular}[c]{@{}c@{}}WG\\ (acc)\end{tabular}} &
  \textbf{\begin{tabular}[c|]{@{}c@{}}MBPP\\ (pass@1)\end{tabular}} &
  \textbf{\begin{tabular}[c|]{@{}c@{}}Avg.\\ (9 tasks)\end{tabular}} \\
  \midrule
\multicolumn{2}{l}{Base model}          & 65.4 & 77.8 & 71.6 & 40.9 & 49.7 & 54.2 & 72.8 & 45.0  & 43.1 & 55.8 \\ \midrule
\multicolumn{2}{l}{\textbf{One-Hot Task E.}}          &  &  &  &  & &  &  &  &  &   \\
\multicolumn{2}{l}{\ours{} (Recon) \Larch} & 76.4	& 89.9	& 89.4	& \greenbox{53.8}	& 92.6	& 85.0	&69.7	& \greenbox{51.2}	& \greenbox{52.6}	& \greenbox{73.4} \\
\multicolumn{2}{l}{\ours{} (Recon) \March} & \greenbox{76.7}	& 89.9	& 89.4	& 53.2	& 92.6	& 85.0	& 69.9	& \greenbox{51.4}	& \greenbox{52.9}	& \greenbox{73.4} \\
\multicolumn{2}{l}{\ours{} (Recon) \Sarch} & 75.2	& 88.8	& 87.4	& 50.9	& 89.1	& 75.6	& \greenbox{83.9}	& \greenbox{58.1}	& 48.1	& 73.0 \\ \midrule
\multicolumn{2}{l}{\textbf{Task Description E.}}          &  &  &  &  & &  &  &  &  &  \\
\multicolumn{2}{l}{\ours{} (Recon) \Larch}& 76.6	& 89.8	& 89.4	& \greenbox{53.9}	& 92.6	& 85.0	& 69.6	& \greenbox{51.2}	& 51.8	& 73.3 \\
\multicolumn{2}{l}{\ours{} (Recon) \March}& 76.5	& 89.9	& 89.4	& \greenbox{53.9}	& 92.5	& 84.9	& \greenbox{70.4}	& \greenbox{51.6}	& \greenbox{52.8}	& \greenbox{73.5} \\
\multicolumn{2}{l}{\ours{} (Recon) \Sarch}& 75.4	& 88.8	& 87.8	& 49.1	& 89.7	& 76.7	& \greenbox{84.2}	& \greenbox{56.9}	& 48.0	& 73.0 \\
\midrule
\multicolumn{2}{l}{Task-specific \lora{}s} & 76.6 & 89.9 & 89.4 & 53.5 & 92.6 & 85.0 & 69.9 & 51.1  & 52.1 & 73.3 \\ \bottomrule
\end{tabular}%
}
\end{table*}
We utilize the SNI dataset \citep{wang2022sni} for training \lora{} adapters. We use a subset of 500 tasks following \citet{bruel2024compress_then_serve}. We use 11 tasks for hold-out validation and removed 10 datasets due to data contamination from the evaluation benchmark tasks, leaving 479 datasets for training.
All samples are in English.
More details of the datasets can be found in \cref{app:datasets}.
For evaluation, we choose 10 widely used benchmarks that collectively cover a variety of LLM capability assessments, e.g., reasoning, math, science, coding, and world knowledge. Specifically, we include the following benchmarks: Arc-challenge (ArcC) and Arc-easy (ArcE) \citep{allenai:arc}, BoolQ \citep{clark2019boolq}, GSM8K \citep{cobbe2021gsm8k}, Hellaswag (HS) \citep{zellers2019hellaswag}, OpenBookQA (OQA) \citep{OpenBookQA2018}, PIQA \citep{Bisk2020piqa}, Winogrande (WG) \citep{ai2:winogrande}, HumanEval (HE) \citep{chen2021humaneval}, and MBPP \citep{austin2021mbpp}. 
\footnote{
The benchmark tasks share some similarities with the training tasks. Specifically, they are mostly multiple-choice question-answering tasks. Also, there are similar and overlapping domains between the two splits. For example, the ARC benchmarks are similar to SNI task \#\href{https://huggingface.co/datasets/Lots-of-LoRAs/task047_miscellaneous_answering_science_questions}{47}.
However, some benchmarks are very different from the training distribution, e.g., MBPP and HumanEval, as the training tasks do not contain any code generation tasks.
}
Task descriptions for the training datasets and the benchmarks are fully generated, as described in \cref{app:gpt_prompt}. When we use a language task embedding as a part of the input, we average \ours{} performance using three descriptions for each benchmark.

\subsection{LoRA Compression}\label{sec:lora_distill}
In this experiment, we aim to investigate whether \ours{} can recover the performance of trained \lora{}s via reconstruction training. For quality control and consistent evaluation, we train a task-specific \lora{} (oracle) on the training split of each benchmark task, collectively forming a library of \lora{}s.
\cref{tab:benchmark-recon} shows the benchmark performance of \ours{} trained by distilling 9 benchmark-specific \lora{}s using either one-hot or natural language task embeddings from \texttt{gte-large-en-v1.5}.
We note that the benchmark tasks are indirectly seen during training by \ours{}, as it learns to distill benchmark-specific \lora{}s.
We can see that \ours{} fully recovers the performance of the oracle adapters with both task embedding types. 
Notably, \ours{} outperforms task-specific \lora{}s on several benchmarks (highlighted in green). We hypothesize that the gain comes from the lossy compression of the target \lora{}s, which acts as a regularization on the already trained \lora{} weights. This effect is most apparent on PIQA and WG benchmarks, where the oracle \lora{} overfits and performs worse than the base model.

\begin{figure}
    \centering
    \includegraphics[width=0.7\linewidth]{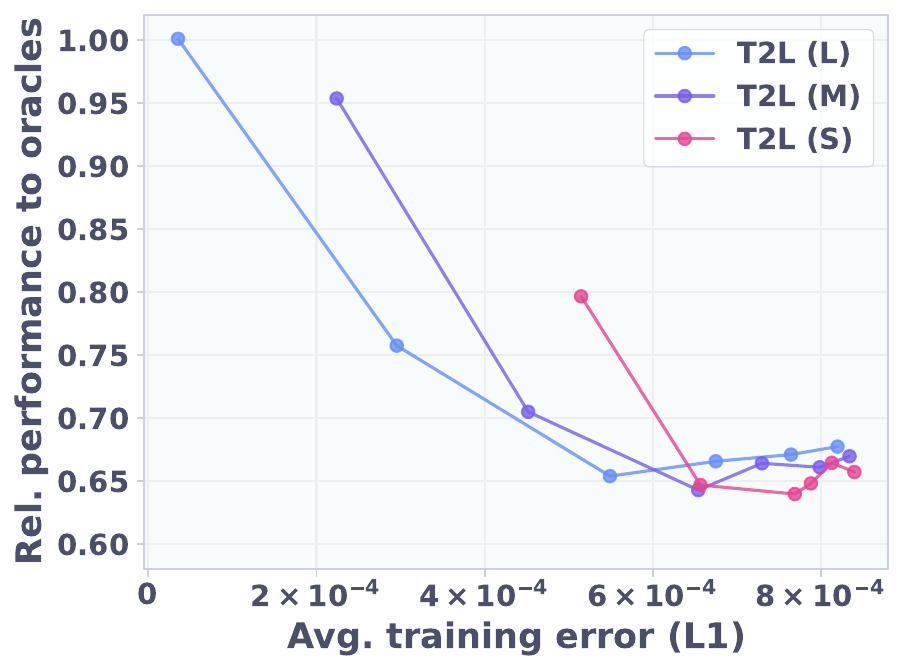}
    \caption{Relative performance and training reconstruction error of \ours{} instances trained with an increasing number of tasks ($\{16,32,64,128,256,479\}$ tasks from left to right).}
    \label{fig:exp_training_perf_vs_err}
    \vspace{-0.5cm}
\end{figure}

\begin{table*}[ht]
\caption{Zero-shot performance on unseen benchmark tasks. SFT-trained \ours{} generates \lora{}s based on unseen task descriptions. Its performance is an average of three generated \lora{}s, each with a different instance of task descriptions. Arrow Routing results are taken from \citet{ostapenko2024towards_modular_llms}. \greenbox{Green highlight} indicates higher performance than that of the benchmark-specific \lora{} adapters. \textbf{Bold numbers} are used when the performance is higher than the multi-task \lora{}.}
\label{tab:zero-shot}
\resizebox{\linewidth}{!}{%
\begin{tabular}{llcccccccc|c|cc|c}
\toprule
\multicolumn{2}{l}{} &
  \textbf{\begin{tabular}[c]{@{}c@{}}ArcC\\ (acc)\end{tabular}} &
  \textbf{\begin{tabular}[c]{@{}c@{}}ArcE\\ (acc)\end{tabular}} &
  \textbf{\begin{tabular}[c]{@{}c@{}}BQ\\ (acc)\end{tabular}} &
  \textbf{\begin{tabular}[c]{@{}c@{}}HS\\ (acc)\end{tabular}} &
  \textbf{\begin{tabular}[c]{@{}c@{}}OQA\\ (acc)\end{tabular}} &
  \textbf{\begin{tabular}[c]{@{}c@{}}PIQA\\ (acc)\end{tabular}} &
  \textbf{\begin{tabular}[c]{@{}c@{}}WG\\ (acc)\end{tabular}} &
  \textbf{\begin{tabular}[c]{@{}c@{}}MBPP\\ (pass@1)\end{tabular}} &
  \textbf{\begin{tabular}[c]{@{}c@{}}Avg.\\ (8 tasks)\end{tabular}} &
  \textbf{\begin{tabular}[c]{@{}c@{}}GSM8K\\ (acc)\end{tabular}} &
  \textbf{\begin{tabular}[c]{@{}c@{}}HE\\ (pass@1)\end{tabular}} &
  \textbf{\begin{tabular}[c]{@{}c@{}}Avg.\\ (10 tasks)\end{tabular}} \\ \midrule
\multicolumn{2}{l}{\textbf{No Test-Time Adaptation}} &      &      &      &      &      &      &      &      &      &      &      &      \\
\multicolumn{2}{l}{\texttt{Mistral-7B-Instruct}}                       & 65.4 & 77.8 & 71.6 & 49.7 & 54.2 & 72.8 & 45.0 & 43.1 & 60.0 & 40.9 & 37.2 & 55.8 \\
\multicolumn{2}{l}{Prepending task desc.} & 72.0 & 85.8 & 67.6 & 58.9 & 63.4 & 77.9 & 59.0 & 41.6 & 65.8 & 40.9  & 39.0 & 60.6 \\
\multicolumn{2}{l}{3-shot ICL} & 72.1 & 85.9 & 71.7 & 59.0 & 66.2 & 76.2 & 58.0 & 42.6 & 66.5 & 40.9  & 37.2 & 61.0 \\
\multicolumn{2}{l}{Average \lora{}}                  & 70.7 & 84.4 & 75.4 & 59.9 & 59.0 & 78.0 & 54.3 & 47.1 & 66.1 & 42.4 & 37.8 & 60.9 \\
\multicolumn{2}{l}{Multi-task \lora{}} & 76.2 & 88.3 & 85.5 & 65.2 & 68.0 & 81.8 & 62.4 & 48.1 & 71.9 & 47.5  & 39.6 & 66.3 \\ \midrule
\multicolumn{2}{l}{\textbf{Zero-Shot Adaptation}}    &      &      &      &      &      &      &      &      &      &      &      &      \\
\multicolumn{2}{l}{Arrow Routing}                    & 60.9 & 86.2 & \textbf{87.6} & \textbf{80.8} & 48.6 & \greenbox{83.0} & \greenbox{68.5} & \textbf{50.2} & 70.7 & N/A  & 28.7 & N/A  \\
\multicolumn{2}{l}{Hyperdecoders (per-instance)} & \textbf{76.6} & \textbf{88.5} & 83.9 & 65.2 & \textbf{76.6} & \greenbox{81.3} & \greenbox{\textbf{64.9}} & \textbf{51.6} & \textbf{73.6} & 43.6 & \textbf{40.9} & \textbf{67.3}  \\\midrule
\multicolumn{2}{l}{\ours{} (SFT) \Sarch}             & 76.0 & \textbf{88.7} & 83.8 & \textbf{68.0} & \textbf{71.6} & \greenbox{82.3} & \greenbox{61.0} & {41.2} & {71.6} & {47.3} & {39.0} & {65.9} \\
\multicolumn{2}{l}{\ours{} (SFT) \March}             & \greenbox{77.2} & \textbf{89.0} & {84.3} & {65.1} & \textbf{76.1} & \greenbox{81.8} & \greenbox{64.0} & \textbf{50.5} & \textbf{73.5} & {45.2} & \textbf{41.3}  & \textbf{67.5} \\
\multicolumn{2}{l}{\ours{} (SFT) \Larch}             & \greenbox{{77.5}} & \textbf{88.9} & {85.0} & \textbf{66.5} & \textbf{75.5} & \greenbox{82.1} & \greenbox{64.2} & \textbf{51.9} & \textbf{73.9} & {45.8} & {39.2} & \textbf{67.7} \\ \midrule
\multicolumn{2}{l}{\textbf{Oracle}}                  &      &      &      &      &      &      &      &      &      &      &      &      \\
\multicolumn{2}{l}{Task-specific \lora{}s}           & 76.6 & 89.9 & 89.4 & 92.6 & 85.0 & 69.9 & 51.1 & 52.1 & 75.8 & 53.5 & N/A  & N/A  \\ \bottomrule
\end{tabular}%
}
\end{table*}
Next, we explore whether \ours{} conditioned on one-hot task vectors can maintain the oracle single-task \lora{}s' performance when using an increasing number of training tasks.
\cref{fig:exp_training_perf_vs_err} shows the performance of one-hot \ours{} on the test splits of a subset of 10 SNI training tasks with varying degrees of final average training L1 reconstruction error.
We train various \ours{} instances for each architecture using $\{16,32,64,128,256,479\}$ training tasks, leading to an effective increase in the training reconstruction error.
Although \ours{} fully recovers the oracles' performance when the reconstruction loss is less than $10^{-4}$, the performance drops as the training error increases. This result suggests that \ours{} learns a lossy compression of the target \lora{}s.
Still, we find that all \ours{} architectures can maintain around $65\%$ of oracles' performance, and the performance does not drop further even at $> 8 \times 10^{-4}$ per-element L1 error. Despite the performance drop, we show that increasing the number of training tasks is beneficial in the SFT setup, increasing zero-shot benchmark performance of \ours{} in \cref{sec:scaling_ds_and_compute}.

\subsection{Zero-Shot LoRA Generation}\label{sec:zeroshot}
Here, we explore whether \ours{} can generate useful \lora{} adapters for unseen tasks. We train \ours{} with SFT on 479 SNI tasks, each with 128 task descriptions. For each data point in a training minibatch, we sample a description from the corresponding dataset in an online fashion.
\cref{tab:zero-shot} shows the zero-shot performance on 10 benchmark tasks.
Here, we present the best model of each variant from our scaling experiment in \cref{sec:scaling_ds_and_compute}.
We observe that a multi-task \lora{} adapter performs well on the benchmarks despite no additional fine-tuning. Still, there is a performance gap between task-specific \lora{}s and MT \lora{}. We observe that SFT-trained \ours{} indeed generates useful \lora{}s, thus improving over the multi-task \lora{} adapter consistently and across benchmarks (indicated by bold numbers). Notably, even though \ours{} cannot fully bridge the performance gap with task-specific \lora{}s, it outperforms the oracles on a subset of tasks (highlighted in green).
We further investigate the generality of our proposed method with different base models including \texttt{Llama} \citep{dubey2024llama} and \texttt{Gemma} \citep{team2024gemma} models in \cref{sec:llama_and_gemma}.
We note that one of the main advantages of T2L is its efficiency. To emphasize T2L’s efficiency, we provide an ad-hoc FLOPs analysis in \cref{app:flops_analysis}.

\section{Ablations and Analyses}
\label{sec:ablations}

\subsection{Increasing Training Compute Proportional to the Number of Training Tasks}\label{sec:scaling_ds_and_compute}
\begin{table*}[th]
    \caption{Performance of SFT-trained \ours{} with varying numbers of training tasks.}
    \label{tab:scaling_ds_and_compute}
    \resizebox{\linewidth}{!}{%
    \begin{tabular}{lcc|cccccccccc|c}
    \toprule
     &
      \textbf{\begin{tabular}[c]{@{}c@{}}Number\\ of tasks\end{tabular}} &
      \textbf{\begin{tabular}[c]{@{}c@{}}Max\\SGD steps\end{tabular}} &
      \textbf{\begin{tabular}[c]{@{}c@{}}ArcC\\ (acc)\end{tabular}} &
      \textbf{\begin{tabular}[c]{@{}c@{}}ArcE\\ (acc)\end{tabular}} &
      \textbf{\begin{tabular}[c]{@{}c@{}}BQ\\ (acc)\end{tabular}} &
      \textbf{\begin{tabular}[c]{@{}c@{}}GSM8K\\ (acc)\end{tabular}} &
      \textbf{\begin{tabular}[c]{@{}c@{}}HS\\ (acc)\end{tabular}} &
      \textbf{\begin{tabular}[c]{@{}c@{}}OQA\\ (acc)\end{tabular}} &
      \textbf{\begin{tabular}[c]{@{}c@{}}PIQA\\ (acc)\end{tabular}} &
      \textbf{\begin{tabular}[c]{@{}c@{}}WG\\ (acc)\end{tabular}} &
      \textbf{\begin{tabular}[c]{@{}c@{}}HE\\ (pass@1)\end{tabular}} &
      \textbf{\begin{tabular}[c]{@{}c@{}}MBPP\\ (pass@1)\end{tabular}} &
      \textbf{Avg.} \\ \midrule
    \multirow{4}{*}{\ours{} (SFT) \Larch} & 479 & 1M & 77.5 &	88.9 &	85.0 &	45.8 &	66.5 &	75.5 &	82.1 &	64.2 &	39.2 &	51.9 &	67.7 \uptriangle \\
                                          & 256 & 640K & 77.3 & 88.1 & 84.3 & 46.0 & 64.5 & 75.7 & 81.9 & 64.0 & 39.8 & 52.1 & 67.4 \uptriangle \\
                                          & 128 & 320K & 76.6 & 88.4 & 85.2 & 46.1 & 67.0 & 74.3 & 81.6 & 55.0  & 38.2 & 45.7 & 65.8 \downtriangle \\
                                          & 64 & 160K & 75.5 & 88.0 & 84.5 & 43.9 & 65.5 & 70.7 & 80.5 & 59.5  & 39.8 & 51.7 & 66.0 \\ \midrule
    \multirow{4}{*}{\ours{} (SFT) \March} & 479 & 1M & 77.2 & 89.0 & 84.3 & 45.2 & 65.1 & 76.1 & 81.8 & 64.0  & 41.3 & 50.5 & 67.5 \uptriangle \\
                                          & 256 & 640K & 75.9 & 89.3 & 85.0 & 47.0 & 65.3 & 73.7 & 81.6 & 63.2  & 39.8 & 48.6 & 66.9 \uptriangle \\
                                          & 128 & 320K & 74.9 & 88.3 & 85.5 & 44.9 & 64.8 & 72.8 & 80.7 & 61.6  & 42.9 & 43.5 & 66.0 \uptriangle \\
                                          & 64 & 160K & 73.6 & 87.7 & 84.5 & 43.2 & 64.6 & 70.5 & 79.9 & 56.0  & 40.7 & 51.4 & 65.2 \\ \midrule
    \multirow{4}{*}{\ours{} (SFT) \Sarch} & 479 & 1M & 77.7 & 88.3 & 85.0 & 46.3 & 65.3 & 73.9 & 82.4 & 61.9  & 34.6 & 36.6 & 65.2 \downtriangle   \\
                                          & 256 & 640K & 76.0 & 88.7 & 83.8 & 47.3 & 68.0 & 71.6 & 82.3 & 61.0  & 39.0 & 41.2 & 65.9 \uptriangle \\
                                          & 128 & 320K & 74.9 & 88.0 & 84.5 & 44.4 & 66.2 & 72.2 & 82.0 & 59.3  & 39.0 & 47.3 & 65.8 \uptriangle \\
                                          & 64 & 160K & 75.4 & 88.4 & 85.0 & 43.1 & 64.8 & 70.7 & 81.5 & 51.6  & 39.4 & 46.7 & 64.7 \\ \bottomrule
    \end{tabular}%
    }
    \end{table*}
In this section, we explore the scalability of \ours{} by varying the training tasks and scale the training budget proportionally to the dataset size on all variants. \cref{tab:scaling_ds_and_compute} shows that, after increasing the number of training tasks and compute budget, \ours{} generally benefits from the additional training tasks. 
However, \Sarch{} does not benefit from extended training with 479 tasks, potentially due to its limited model capacity.
We additionally investigate the effect of the task diversity on the robustness of \ours{} by training on more tasks without scaling the training budget in \cref{sec:scale_num_ds}. We find that it is crucial to scale the compute budget according to the number of training tasks. For instance, \March{} with scaled compute budget improves over training runs with a fixed budget when using 256 or more training tasks.

\subsection{Task Embedding Models}\label{sec:task_emb_models}
\begin{table}[thb]
    \centering
    \caption{Zero-shot benchmark performance of SFT \ours{} trained on 128 tasks using different text embedding models.}
    \label{tab:vary_emb_model}
    \scalebox{0.8}{%
    \begin{tabular}{lcccccc}
    \toprule
                           & \multicolumn{3}{c}{\texttt{gte}} & \multicolumn{3}{c}{\texttt{Mistral}} \\ \cmidrule(lr){2-4} \cmidrule(lr){5-7}
    \multirow{2}{*}{\begin{tabular}[c]{@{}l@{}}Avg. Benchmark\\ performance\end{tabular}}                       & \Sarch{}   & \March{}  & \Larch{}  & \Sarch{}  & \March{}  & \Larch{} \\
                          & {65.8}      & {66.0}      & {65.8}   & {64.7} & {66.2} & {66.0} \\ \midrule
    \textbf{Avg.}         & \multicolumn{3}{c}{\textbf{65.9}} & \multicolumn{3}{c}{\textbf{65.6}} \\ \bottomrule
    \end{tabular}%
    }
\end{table}
\cref{tab:vary_emb_model} shows the zero-shot benchmark performance with two different embedding models: \texttt{gte-large-en-v1.5} and \texttt{Mistral-7B-Instruct}. For the \texttt{gte} model, we extract a task description by using the activation of the \texttt{CLS} token in the last layer as the model is a bidirectional model. For \texttt{Mistral}, we use the activation of the last token in the sequence to represent a given description \citep{llm2vec}. 
\cref{tab:vary_emb_model} shows the results with the two embedding models used for \ours{} SFT training on 128 tasks.
Both embedding models yield \ours{} instances with comparable generalization capability, suggesting \ours{}'s robustness to task description embedding methods. 

\subsection{Varying Task Descriptions}\label{sec:vary_task_desc}
\begin{figure*}[ht]
    \centering
    \includegraphics[page=15,trim=2.5cm 0cm 2.5cm 0cm,clip,width=0.7\linewidth]{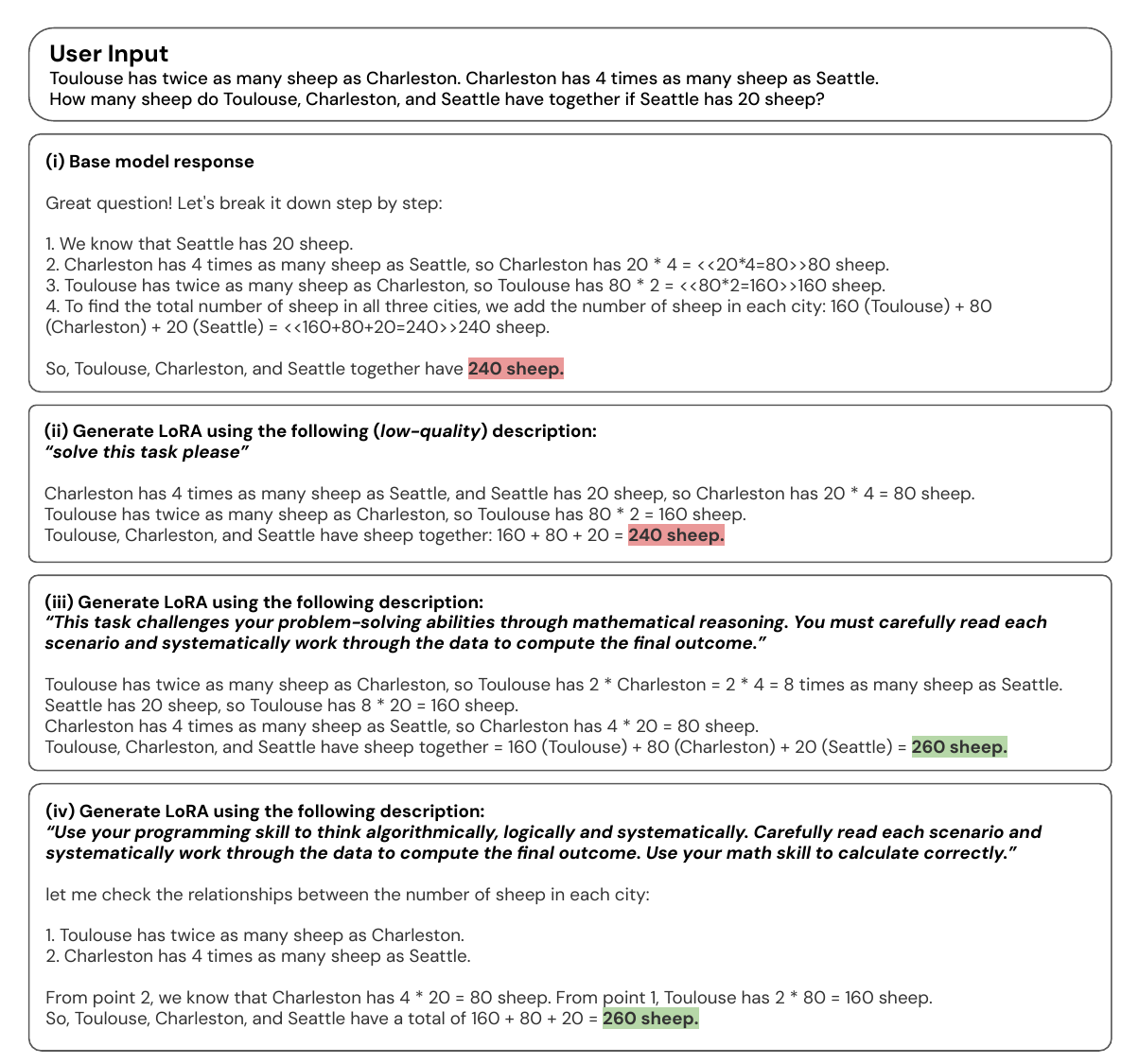}
    \caption{Qualitative examples of responses from applying LoRA generated by \ours{} to the \textit{Mistral-7B-Instruct} base model on a GSM8K problem instance. \textbf{(i)} The response from the base model is incorrect. \textbf{(ii)} Applying a LoRA generated from a \emph{low-quality} task description does not make the model output the correct response. \textbf{(iii, iv)} Descriptions that are aligned with the problem lead to generated LoRAs that steer the base model to output correct responses. Descriptions from \textbf{(iii)} and \textbf{(iv)} influence the model to generate different reasoning paths, highlighting the steerability of \ours{}.
    }
    \label{fig:qualitative}
\end{figure*}

\begin{table}[thb]
    \centering
    \caption{\ours{} trained via reconstruction on 9 tasks performs well when given aligned task descriptions. Unaligned descriptions produce lower benchmark performance.}
    \label{tab:vary_task_desc}
    \scalebox{0.8}{%
    \begin{tabular}{lcccc}
    \toprule
                                          & \multicolumn{2}{c}{\textbf{Aligned}} & \multicolumn{2}{c}{\textbf{Unaligned}} \\ \cmidrule(lr){2-3} \cmidrule(lr){4-5}
                                          & \textbf{Train} & \textbf{Eval} & \textbf{Train (random)} & \textbf{Random strings} \\ \midrule
    \multicolumn{1}{l}{\ours{} \Larch{}} & 73.3           & 73.6          & 49.1                    & 68.2                    \\
    \multicolumn{1}{l}{\ours{} \March{}} & 73.5           & 70.2          & 49.5                    & 68.5                    \\
    \multicolumn{1}{l}{\ours{} \Sarch{}} & 73.0           & 72.9          & 55.7                    & 53.9                    \\ \midrule 
    \textbf{Avg.}                        & \textbf{73.3}           & \textbf{72.2}          & \textbf{51.4}                    & \textbf{63.5}                    \\ \bottomrule
    \end{tabular}%
    }
\end{table}
We investigate the impact of task descriptions on the performance of generated \lora{}s using four types of descriptions:
\begin{itemize}
    \itemsep=0.5em
    \item \textbf{Train:} Training descriptions of corresponding tasks.
    \item \textbf{Eval:} Unseen descriptions of corresponding tasks.
    \item \textbf{Random strings:} Random literal strings.
    \item \textbf{Train (random):} Training descriptions randomly sampled from other tasks.
\end{itemize}
For each description type, we use the \texttt{gte-large-en-v1.5} embedding and report the average performance using three descriptions. The four types can be grouped into two categories based on the alignment between the descriptions and the tasks: aligned (\textbf{Train}, \textbf{Eval}) and unaligned (\textbf{Train (random)} and \textbf{Random strings}).
Note that we use reconstruction-trained \ours{} in this experiment. That is, the hypernetwork has seen training descriptions of the benchmarks during training.
We observe a performance gap between the two description categories. Specifically, training and evaluation descriptions generate the best performing \lora{}s, matching the performance of oracle \lora{}s, despite the evaluation descriptions being unseen. These results suggest that \ours{} is robust to changes in the task description as long as the descriptions are aligned with the task. On the other hand, if the descriptions are not aligned with the task at hand, the generated \lora{}s will not perform as well, as indicated by the performance of the unaligned group.
We believe that using an LLM for adjusting the description alignment could effectively sidestep this failure case of T2L.
Additionally, we provide a qualitative result demonstrating steerability and an unsuccessful example of \ours{} in \cref{fig:qualitative}. Importantly, the last two examples in \cref{fig:qualitative} \textbf{(iii, iv)} are both correct but have different answer styles thanks to different descriptions. We remark that Hyperdecoders \citep{ivison2022hyperdecoders} cannot exhibit such steerability as it uses the problem instance as the input to the hypernetwork.

\subsection{Training Schemes}\label{subsec:recon_vs_sft_zeroshot}
\begin{table}
    \centering
    \caption{Reconstruction vs SFT training scheme.}
    \label{tab:zero_shot_recon_vs_sft}
    \scalebox{0.8}{%
    \begin{tabular}{lcccccc}
    \toprule
                           & \multicolumn{3}{c}{\textbf{Recon}} & \multicolumn{3}{c}{\textbf{SFT}} \\ \cmidrule(lr){2-4} \cmidrule(lr){5-7}
    \multirow{2}{*}{\begin{tabular}[c]{@{}l@{}}Benchmark\\ performance\end{tabular}}                       & \Sarch{}   & \March{}  & \Larch{}  & \Sarch{}  & \March{}  & \Larch{} \\
                          & 61.8       & 61.7      & 62.0      & {64.8}      & {66.5}      & {67.5} \\ \midrule
     \textbf{Avg.}        & \multicolumn{3}{c}{$\textbf{61.8}$} & \multicolumn{3}{c}{$\textbf{66.3}$} \\ \bottomrule
    \end{tabular}%
    }
\end{table}
In this section, we investigate the zero-shot performance of SFT-trained and reconstruction-trained \ours{}. All model instances are trained with roughly equal wall-clock time of 10 hours (see \cref{app:reproducibility} for details). From \cref{tab:zero_shot_recon_vs_sft}, we can see a clear performance gap between reconstruction and SFT training schemes. Specifically, SFT produces \ours{} instances that perform significantly better than those trained via reconstruction ($66.3$ vs $61.83$ benchmark performance averaged over model architectures).
We attribute the performance difference to the library of \lora{}s needed for reconstruction training. For reconstruction-trained \ours{} to generalize, the target \lora{} adapters of similar tasks should be clustered in some latent manifold. In contrast, SFT training does not need pre-trained task-specific \lora{} adapters, thus sidestepping this challenge via end-to-end learning. In \cref{subsec:lora_of_similar_tasks}, we show that pre-trained adapters for similar tasks do not live nearby in the weight space, supporting our claim of a potential problem when reconstructing pre-trained \lora{}s.

\subsection{Visualization of \ours{} Activations}\label{subsec:umap_visualization}
\begin{figure*}[th]
    \vspace{-0.5cm}
    \centering
    \subfloat{
        \includegraphics[trim={0cm 0cm 6cm 0cm},clip,width=0.35\linewidth]{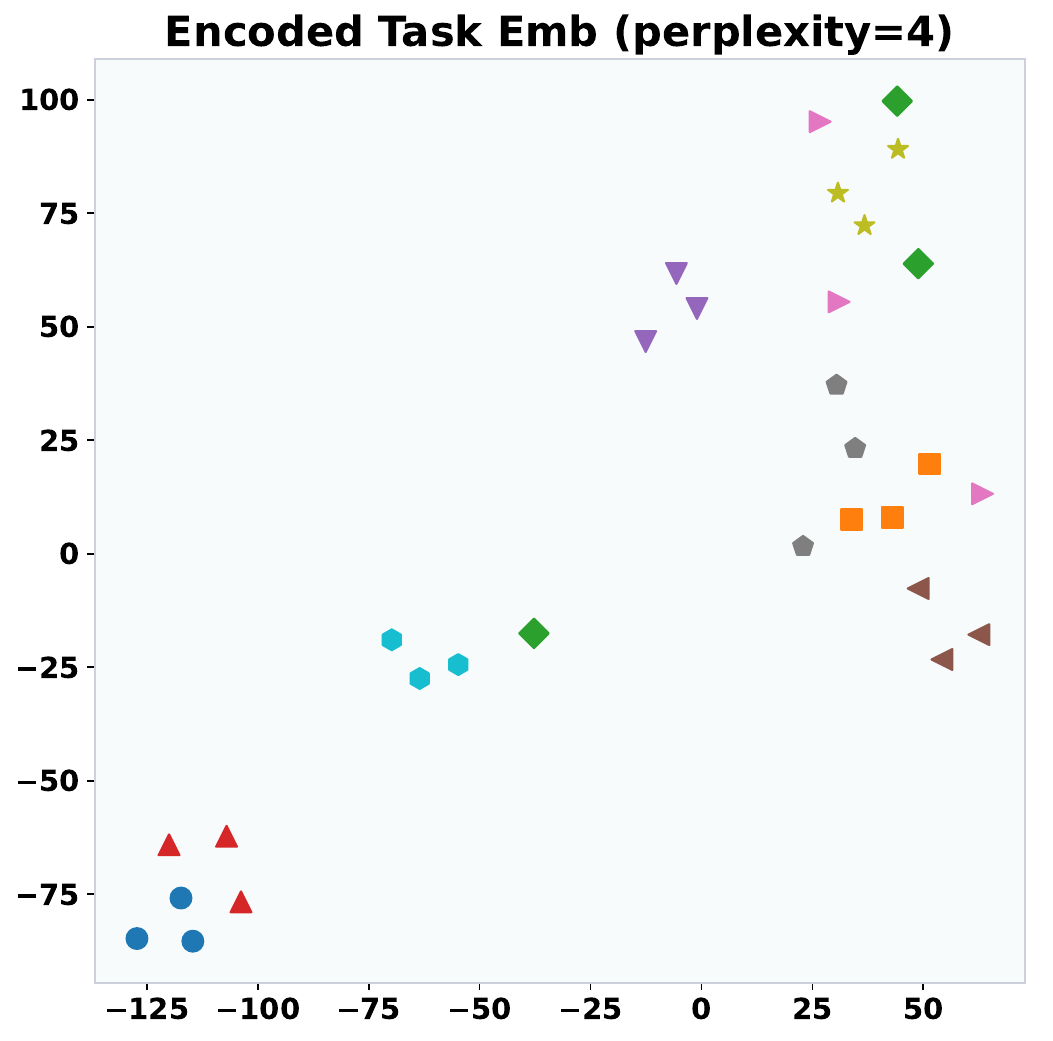}
    }
    \subfloat{
        \includegraphics[trim={0cm 0cm 6cm 0cm},clip,width=0.35\linewidth]{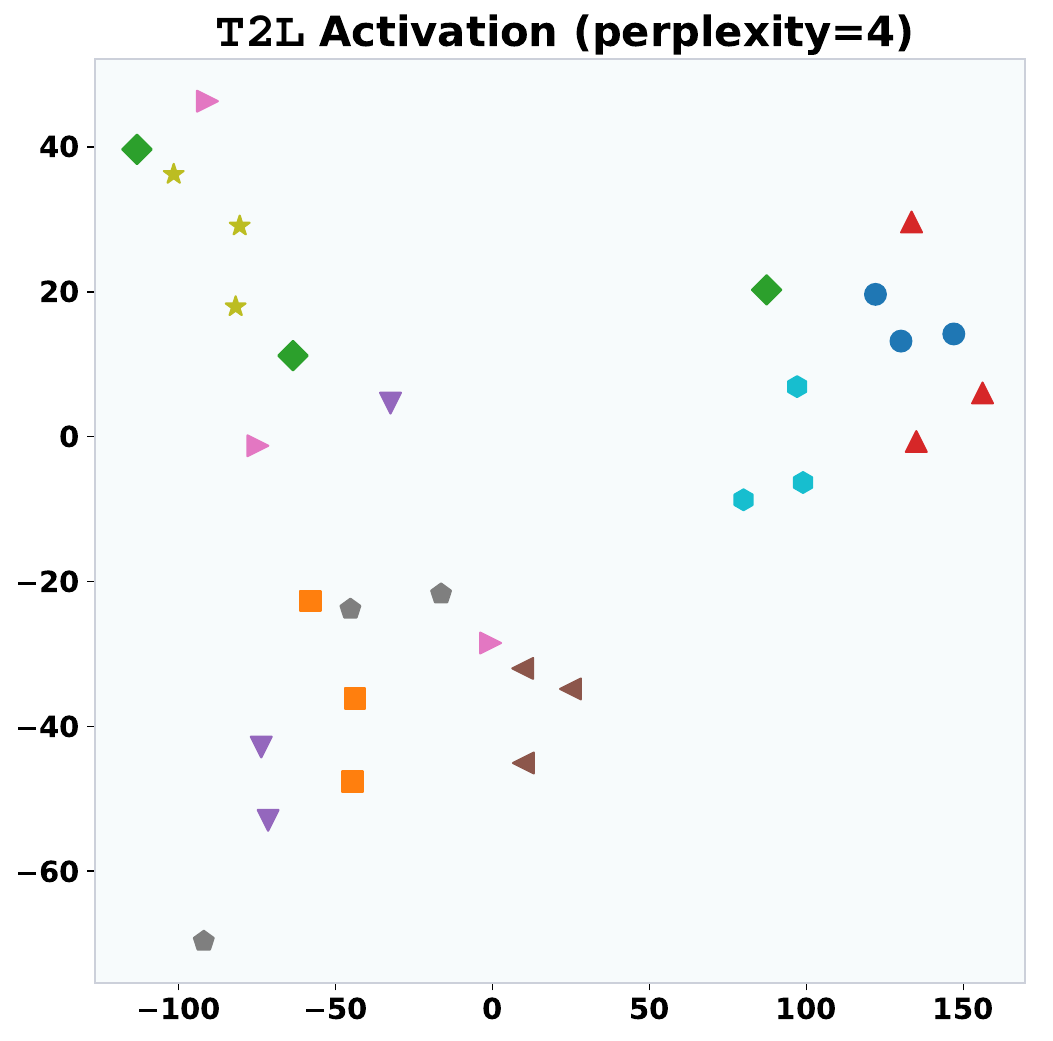}
    }
    \subfloat{
        \includegraphics[trim={17.8cm 0cm 0cm 9cm},clip,width=0.15\linewidth]{figs/tsne/HyperLoRA_Activation_perplexity=4.pdf}
    }
    \caption{2D t-SNE projection of activations of \ours{}'s task encoder (left) and activations of the last MLP block (right) grouped by benchmark tasks (represented by colors). We probe \ours{} with unseen three task descriptions per benchmark. We can see activations clustering in both plots, indicating that \ours{} indeed learns to generate \lora{}s tailored to specific tasks.}
    \label{fig:tsne_activation}
    \vspace{-0.2cm}
\end{figure*}
Next, we aim to understand \ours{} further and see whether it generates task-specific \lora{} adapters for unseen tasks with unseen descriptions. We probe SFT \ours{} \March{} trained on 256 training tasks in the zero-shot evaluation setting.  The model is probed on all the benchmark tasks, each with three unseen descriptions. \cref{fig:tsne_activation} shows the 2D t-SNE projection of \ours{}'s task encoder activations and the outputs of the last MLP block.
We can see clear clustering in both projection plots based on the tasks (colors and shapes). \ours{} generates different adapters for different tasks, confirming that \ours{} indeed performs task-specific adaptation `on the fly'.
Moreover, similar tasks, e.g., MBPP and HumanEval, are clustered together in both plots, suggesting that SFT-trained \ours{} produces similar adapters for semantically similar tasks.

\section{Related Work}

\textbf{Hypernetworks for Adaptation}: Hypernetworks \citep{ha2016hypernetworks} provide a general indirect encoding method for neural network weights. They have been applied to different architectures \citep[e.g., in attention, ][]{schug2024attention} and training paradigms \citep[e.g., in continual learning, ][]{von2019continual}. Here, we focus on generating low-rank adapters using natural language instructions. Previous work \citep{mahabadi2021parameter, he2022hyperprompt, ortiz2024hyperloader} considers hypernetworks for LLM adaptation in a multi-task context but only uses learned task identifiers instead of natural language for adaptation. Thus, these approaches do not enable task-wise zero-shot generalization.

\textbf{Hypernetworks for Zero-Shot LLM Adaptation:}
\citet{xiao-etal-2023-hyperlora-dialect} explore the use of hypernetworks on a limited set of English dialects; they only consider five dialects, one of which is unseen. Furthermore, the hypernetwork relies on an expert-based transformation of the dialects, limiting the possibility for generalization.
\citet{mu2024gisting} propose Gisting, a method that learns to compress an in-context task description to prefix tokens, allowing the language model to follow instructions with fewer tokens. However, Gisting is limited to prefix tokens---only influencing the attention matrices of the base model. Thus, prefix tokens are less flexible compared to \lora{}s that can modify different parts of LLMs, e.g., attention blocks.
Hyperdecoders \citep{ivison2022hyperdecoders} is a hypernetwork that generates adapters on the fly based on the input sequence. While per-sequence adaptation is desirable for benchmark evaluation---where the LLM should always output the correct answer---we argue that description-based adaptation gives more control to users since they can steer the LLM in creative ways based on user-generated descriptions (see \cref{fig:qualitative}). Furthermore, the generated adapters cannot be efficiently fused into the base model, leading to significant overhead for each query.

Closely related to our work are HyperTuning \citep{phang2023hypertuning}, HNET-LM \citep{deb-etal-2022-hnet-lm}, and HINT \citep{ivison2023hint}. Differing from prior work that heavily focuses on pre-trained encoder-decoder models, e.g., T5 \citep{raffel2020t5} or BART \citep{lewis2019bart}, we use frontier instruction fine-tuned models as the base models, i.e., \texttt{Mistral}, \texttt{Llama}, \texttt{Gemma}. Also, prior work typically relies on initializing a part of their hypernetworks from the base model (e.g., tying task encoder's weights to the base model) to achieve good performance or stable training as opposed to ours that are task-embedder agnostic and can freely change the task embedding model (\cref{sec:task_emb_models}). Additionally, our work utilizes generated descriptions instead of the ones provided by the SNI dataset.
In \cref{sec:description_source}, we show that using generated descriptions increase the performance of \ours{} considerably.
Overall, our work improves upon prior work in several ways, including achieving task-wise zero-shot generalization on various frontier instruction-tuned language models, simpler and more general hypernetwork input requirements, investigation of training regimes, and more comprehensive experiments, ablations, and analyses.

Concurrent to our work, \citet{Lv2024hyperlora-few-shot} propose a similar approach that utilizes a hypernetwork to generate \lora{} adapters at inference time. However, their hypernetwork assumes that the context vector provided to the hypernetwork contains few-shot examples. In contrast, \ours{} only assumes a task description, which users can produce themselves within seconds.

\section{Discussion and Limitations}

\textbf{Discussion}.
We rely on generated descriptions from \texttt{GPT-4o mini} to ensure high-quality and consistent task descriptions. It is plausible that when \ours{} is deployed in real-world scenarios, users might not input high-quality descriptions, which could cause performance degradation of generated adapters.
Our results have primarily focused on LLM adaptation.
However, \ours{} can be directly applied to other LLMs or to adapt vision language models.
Finally, the potential for \ours{} trained on a smaller base model to transfer effectively to larger models within the same architecture class remains an open area for exploration.

\textbf{Limitations}.
We only consider \lora{} as the output space of the hypernetwork. We believe there are more efficient ways to modulate LLMs given a text description, e.g., directly modulating the activations of the base model.
Also, we believe the compression achieved by \ours{} can be further optimized using well-designed inductive biases.
Finally, although \ours{} exhibits robustness and signs of scalability, it still does not reach the benchmark performance of task-specific \lora{}s in a zero-shot manner. Achieving such potent zero-shot adaption is still one of the biggest challenges for \ours{}.

\section*{Acknowledgment}
We thank David Ha for suggesting \emph{``Text-to-LoRA''} as the title of the paper. We thank anonymous reviewers for their constructive feedback, which we incorporate to improve the quality of the paper.

\section*{Impact Statement}
This paper introduces Text-to-LoRA (\ours{}), a novel approach that significantly lowers the barrier to adapting large foundation models for specific tasks. Traditionally, customizing models like LLMs requires resource-intensive fine-tuning on specific datasets for each new application, limiting accessibility and slowing down deployment. \ours{} overcomes this by training a hypernetwork to generate task-specific Low-Rank Adapters (LoRAs) instantly, using only a natural language description of the target task as input. This eliminates the need for per-task fine-tuning datasets and lengthy optimization processes, enabling rapid, on-the-fly adaptation with minimal computational overhead during inference, thereby making powerful model customization more accessible.

The broader impact of \ours{} lies in its potential to democratize the specialization of powerful AI systems by enabling adaptation through intuitive text instructions. While \ours{} demonstrates effective compression and promising zero-shot generalization to unseen tasks similar to those encountered during training, potential pitfalls exist that warrant consideration. Its performance is notably sensitive to the quality and clarity of the natural language task descriptions; poorly phrased or misaligned instructions could lead to suboptimal or incorrect adaptations, potentially hindering reliability in real-world user scenarios. Furthermore, while \ours{} significantly advances instant adaptation, its generalization capability to task types fundamentally different from its training distribution (e.g., beyond the SNI-derived benchmarks) needs further investigation, and it may not yet fully match the performance ceiling of adapters meticulously fine-tuned on extensive, high-quality datasets for highly complex or specialized domains.

\bibliography{main}
\bibliographystyle{icml2025}

\newpage
\appendix
\onecolumn

\section{Generalization to \texttt{Llama} and \texttt{Gemma} Models}\label{sec:llama_and_gemma}
\begin{table}[th]
\caption{Zero-shot performance with \texttt{Llama-3.1-8B-Instruct} as the base language model.}
\label{tab:zero-shot-llama}
\resizebox{\columnwidth}{!}{%
\begin{tabular}{llcccccccccc|c}
\toprule
\multicolumn{2}{l}{} &  \textbf{\begin{tabular}[c]{@{}c@{}}ArcC\\ (acc)\end{tabular}} &
  \textbf{\begin{tabular}[c]{@{}c@{}}ArcE\\ (acc)\end{tabular}} &
  \textbf{\begin{tabular}[c]{@{}c@{}}BQ\\ (acc)\end{tabular}} &
  \textbf{\begin{tabular}[c]{@{}c@{}}GSM8K\\ (acc)\end{tabular}} &
  \textbf{\begin{tabular}[c]{@{}c@{}}HS\\ (acc)\end{tabular}} &
  \textbf{\begin{tabular}[c]{@{}c@{}}OQA\\ (acc)\end{tabular}} &
  \textbf{\begin{tabular}[c]{@{}c@{}}PIQA\\ (acc)\end{tabular}} &
  \textbf{\begin{tabular}[c]{@{}c@{}}WG\\ (acc)\end{tabular}} &
  \textbf{\begin{tabular}[c]{@{}c@{}}HE\\ (pass@1)\end{tabular}} &
  \textbf{\begin{tabular}[c]{@{}c@{}}MBPP\\ (pass@1)\end{tabular}} &
  \textbf{Avg.} \\ \midrule
\multicolumn{2}{l}{\texttt{Llama-3.1-8B-Instruct}} & {73.3} &	{90.6} &	{80.4} &	{75.7} &	{66.6} &	{75.4} &	{79.8} &	{55.3} &	{	66.5} &	{68.7} &	{73.2}  \\
\multicolumn{2}{l}{3-shot ICL} & {80.7} &	{91.9} &	{80.0} &	{75.7} &	{59.3} &	{77.6} &	{80.9} &	{61.3} &	{	66.5} &	{70.4} &	{74.4}  \\
\multicolumn{2}{l}{Prepending task desc.} & {80.2} &	{92.5} &	{79.9} &	{75.7} &	{69.8} &	{78.4} &	{81.7} &	{62.4} &	{	68.3} &	{70.2} &	{75.9}   \\
\multicolumn{2}{l}{Multi-task \lora{}} & {82.0} &	{92.8} &	{83.3} &	{77.6} &	{70.8} &	{\textbf{81.8}} &	{\textbf{83.8}} &	{\textbf{60.3}} &	{	63.4} &	{69.4} &	{76.5}  \\ \midrule
\multicolumn{2}{l}{\ours{} (SFT) \Larch} & { \textbf{82.4}} &	{\textbf{92.9}} &	{\textbf{84.4}} &	{\textbf{79.1}} &	{\textbf{72.8}} &	{\textbf{81.8}} &	{81.2} &	{60.0} &	{\textbf{64.6}} &	{\textbf{69.9}} &	{\textbf{76.9}}    \\ \bottomrule
\end{tabular}%
}
\end{table}
\begin{table}[th]
\caption{Zero-shot performance with \texttt{Gemma-2-2B-Instruct} as the base language model.}
\label{tab:zero-shot-gemma}
\resizebox{\columnwidth}{!}{%
\begin{tabular}{llcccccccccc|c}
\toprule
\multicolumn{2}{l}{} &  \textbf{\begin{tabular}[c]{@{}c@{}}ArcC\\ (acc)\end{tabular}} &
  \textbf{\begin{tabular}[c]{@{}c@{}}ArcE\\ (acc)\end{tabular}} &
  \textbf{\begin{tabular}[c]{@{}c@{}}BQ\\ (acc)\end{tabular}} &
  \textbf{\begin{tabular}[c]{@{}c@{}}GSM8K\\ (acc)\end{tabular}} &
  \textbf{\begin{tabular}[c]{@{}c@{}}HS\\ (acc)\end{tabular}} &
  \textbf{\begin{tabular}[c]{@{}c@{}}OQA\\ (acc)\end{tabular}} &
  \textbf{\begin{tabular}[c]{@{}c@{}}PIQA\\ (acc)\end{tabular}} &
  \textbf{\begin{tabular}[c]{@{}c@{}}WG\\ (acc)\end{tabular}} &
  \textbf{\begin{tabular}[c]{@{}c@{}}HE\\ (pass@1)\end{tabular}} &
  \textbf{\begin{tabular}[c]{@{}c@{}}MBPP\\ (pass@1)\end{tabular}} &
  \textbf{Avg.} \\ \midrule
\multicolumn{2}{l}{\texttt{Gemma-2-2B-Instruct}}                       & 73.7 & \textbf{89.9} & 81.0 & 55.6 & 55.2 & 71.0 & 71.0 & 53.8 & \textbf{43.9} & 12.3 & 60.7 \\
\multicolumn{2}{l}{3-shot ICL} &  72.4 & 88.9 & \textbf{82.5} & 55.6 & 55.7 & 72.6 & 67.6 & 53.7 & \textbf{43.9} & 43.1 & 63.6 \\
\multicolumn{2}{l}{Prepending task desc. w/ ICL} & 72.4 & 88.9 & \textbf{82.5} & 55.6 & 55.7 & 72.6 & 67.6 & 53.7 & \textbf{43.9} & 43.1 & 63.6 \\
\multicolumn{2}{l}{Multi-task \lora{} w/ ICL} & 73.5 & 89.4 & 81.6 & \textbf{57.2} & 59.5 & \textbf{74.6} & 69.4 & 58.1 & 39.0 & 50.4 & 65.2 \\ \midrule
\multicolumn{2}{l}{\ours{} (SFT) \Larch{} w/ ICL} & \textbf{74.0} & 89.8 & 81.8 & 55.1 & \textbf{62.5} & 73.9 & \textbf{75.2} & \textbf{58.7} & 41.5 & \textbf{51.5} & \textbf{66.4} \\ \bottomrule
\end{tabular}%
}
\end{table}

In this section, we explore the generality of our proposed architectures to different model families and sizes.
\cref{tab:zero-shot-llama,tab:zero-shot-gemma} show the benchmark performance of \ours{} \Larch{} compared to various baselines using \texttt{Llama-3.1-8B-Instruct} and \texttt{Gemma-2-2B-Instruct} as the base models, respectively. With \texttt{Gemma} base model, we utilize ICL for all approaches as it drastically improves the performance on the MBPP benchmark.
We see that \ours{} consistently outperforms the baselines across all tested models with varying model sizes and architectures. We note that \ours{} are trained with the same set of hyperparameters across base models.

\section{Training Description Sources}\label{sec:description_source}
\begin{table}[th]
\caption{Performance of SFT-trained \ours{} with two different training description sources.}
\label{tab:trainng_desc_source}
\resizebox{\columnwidth}{!}{%
\begin{tabular}{lcccccccccc|c}
\toprule &
  \textbf{\begin{tabular}[c]{@{}c@{}}ArcC\\ (acc)\end{tabular}} &
  \textbf{\begin{tabular}[c]{@{}c@{}}ArcE\\ (acc)\end{tabular}} &
  \textbf{\begin{tabular}[c]{@{}c@{}}BQ\\ (acc)\end{tabular}} &
  \textbf{\begin{tabular}[c]{@{}c@{}}GSM8K\\ (acc)\end{tabular}} &
  \textbf{\begin{tabular}[c]{@{}c@{}}HS\\ (acc)\end{tabular}} &
  \textbf{\begin{tabular}[c]{@{}c@{}}OQA\\ (acc)\end{tabular}} &
  \textbf{\begin{tabular}[c]{@{}c@{}}PIQA\\ (acc)\end{tabular}} &
  \textbf{\begin{tabular}[c]{@{}c@{}}WG\\ (acc)\end{tabular}} &
  \textbf{\begin{tabular}[c]{@{}c@{}}HE\\ (pass@1)\end{tabular}} &
  \textbf{\begin{tabular}[c]{@{}c@{}}MBPP\\ (pass@1)\end{tabular}} &
  \textbf{Avg.} \\ \midrule
\ours{} (SFT) \Larch & 77.5 &	88.9 &	85.0 &	45.8 &	66.5 &	75.5 &	82.1 &	64.2 &	39.2 &	51.9 &	67.7 \\
\ours{} (SFT) \Larch{} w/ SNI def.  & 75.3 & 87.4 & 85.0 & 45.9 & 63.6 & 73.5 & 80.9 & 61.8 & 38.2 & 53.8 & 66.5 \\ \bottomrule
\end{tabular}%
}
\end{table}
In this experiment, we explore the impact of the sources of the training task descriptions: SNI and chatGPT (\cref{app:gpt_prompt}) \cref{tab:trainng_desc_source} shows that using task definitions provided by the SNI datasets reduces the zero-shot benchmark performance of \ours{}. As the SNI datasets are crowd-sourced, we hypothesized that the task descriptions might have inconsistent template or varied levels of details. Thus, it is harder for \ours{} to learn and generalize.

\section{Scaling the Number of Training Tasks with Fixed Compute}\label{sec:scale_num_ds}
\begin{table}[th]
\caption{Benchmark performance of SFT-trained \ours{} with varying numbers of training tasks. We show results with $\{64,128,256,479\}$ tasks. \uptriangle{} (\downtriangle) indicates increased (decreased) performance compared to the previous increment in the number of training tasks and training budget.}
\label{tab:scaling}
\resizebox{\columnwidth}{!}{%
\begin{tabular}{lc|cccccccccc|c}
\toprule
 &
  \textbf{\begin{tabular}[c]{@{}l@{}}Number\\ of tasks\end{tabular}} &
  \textbf{\begin{tabular}[c]{@{}c@{}}ArcC\\ (acc)\end{tabular}} &
  \textbf{\begin{tabular}[c]{@{}c@{}}ArcE\\ (acc)\end{tabular}} &
  \textbf{\begin{tabular}[c]{@{}c@{}}BQ\\ (acc)\end{tabular}} &
  \textbf{\begin{tabular}[c]{@{}c@{}}GSM8K\\ (acc)\end{tabular}} &
  \textbf{\begin{tabular}[c]{@{}c@{}}HS\\ (acc)\end{tabular}} &
  \textbf{\begin{tabular}[c]{@{}c@{}}OQA\\ (acc)\end{tabular}} &
  \textbf{\begin{tabular}[c]{@{}c@{}}PIQA\\ (acc)\end{tabular}} &
  \textbf{\begin{tabular}[c]{@{}c@{}}WG\\ (acc)\end{tabular}} &
  \textbf{\begin{tabular}[c]{@{}c@{}}HE\\ (pass@1)\end{tabular}} &
  \textbf{\begin{tabular}[c]{@{}c@{}}MBPP\\ (pass@1)\end{tabular}} &
  \textbf{Avg.} \\ \midrule
\multirow{4}{*}{\ours{} (SFT) \Larch} & 479 & 77.2 & 89.0 & 85.0 & 46.3 & 66.5 & 73.6 & 82.6 & 61.8 & 39.2 & 44.3 & 66.6  \downtriangle \\
                                      & 256 & 76.6 & 89.1 & 84.8 & 47.0 & 67.7 & 73.5 & 82.8 & 62.4 & 39.6 & 51.0 & 67.5  \uptriangle \\
                                      & 128 & 76.2 &	89.0 &	85.3 &	46.2 &	67.9 &	71.7 &	82.6 &	59.9 &	40.5 &	51.3 &	67.0  \uptriangle \\
                                      & 64 & 75.5 &	88.0 &	84.5 &	43.9 &	65.5 &	70.7 &	80.5 &	59.5 &	39.8 &	51.7 &	66.0                 \\ \midrule
\multirow{4}{*}{\ours{} (SFT) \March} & 479 & 77.5 & 89.0 & 85.0 & 45.8 & 66.5 & 71.9 & 82.1 & 61.4 & 41.3 & 50.1 & 67.1  \uptriangle \\
                                      & 256 & 76.1 & 88.2 & 85.3 & 45.4 & 65.6 & 72.7 & 81.7 & 62.3 & 36.8 & 50.6 & 66.5  \uptriangle \\
                                      & 128 & 75.5 & 87.8 & 85.3 & 46.1 & 66.6 & 71.6 & 81.7 & 62.2 & 39.8 & 44.9 & 66.1  \uptriangle \\ 
                                      & 64 & 73.6 & 87.7 & 84.5 & 43.2 & 64.6 & 70.5 & 79.9 & 56.0 & 40.7 & 51.4 & 65.2                 \\ \midrule
\multirow{4}{*}{\ours{} (SFT) \Sarch} & 479 & 75.8 & 88.5 & 83.9 & 45.6 & 64.2 & 71.9 & 82.3 & 61.5 & 36.2 & 45.0 & 65.5  \uptriangle   \\
                                      & 256 & 76.1 & 88.4 & 83.0 & 47.3 & 65.0 & 71.7 & 82.5 & 58.1 & 36.2 & 39.1 & 64.8  \uptriangle \\
                                      & 128 & 75.6 & 87.7 & 84.9 & 46.5 & 65.7 & 72.7 & 81.0 & 59.6 & 39.0 & 28.1 & 64.1  \downtriangle \\ 
                                      & 64 & 75.4 & 88.4 & 85.0 & 43.1 & 64.8 & 70.7 & 81.5 & 51.6 & 39.4 & 46.7 & 64.7                 \\ \bottomrule
\end{tabular}%
}
\end{table}
We study the impact of the number of training tasks on the zero-shot benchmark performance of \ours{} in the SFT setting, where all \ours{} instances are trained for roughly the same number of gradient steps (see details in \cref{app:reproducibility}). Overall, we find that increasing the number of training tasks improves the average zero-shot benchmark performance of the hypernetwork (\cref{fig:conceptual,tab:scaling}).
This result hints at the plausible scalability of \ours{} and positive transfer between tasks.

\section{LoRAs of Similar Tasks}\label{subsec:lora_of_similar_tasks}
\begin{figure}[th]
    \centering
    \includegraphics[width=0.32\linewidth]{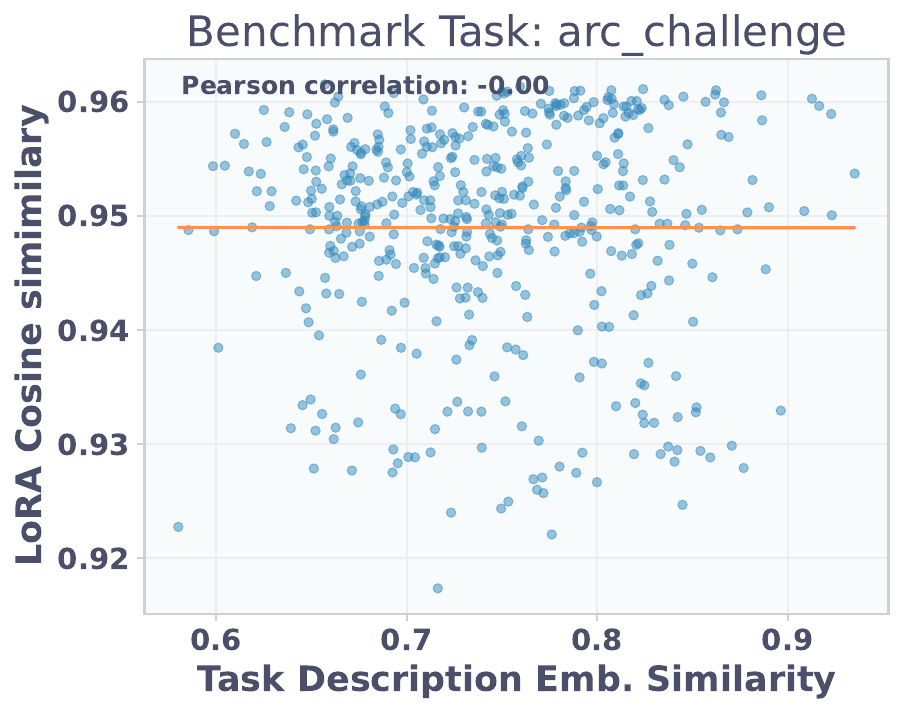}
    \includegraphics[width=0.32\linewidth]{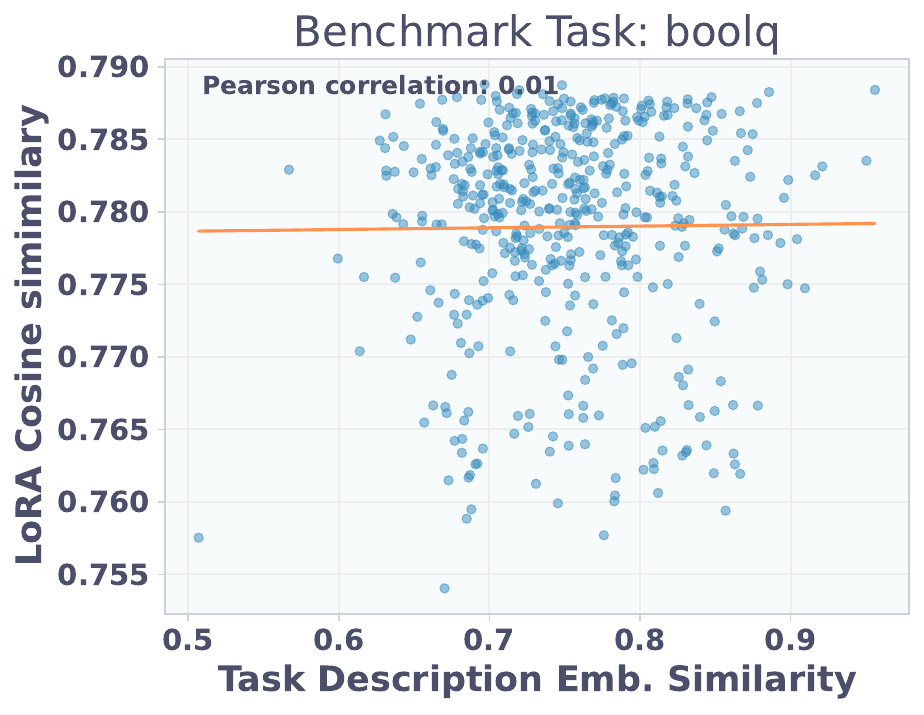}
    \includegraphics[width=0.32\linewidth]{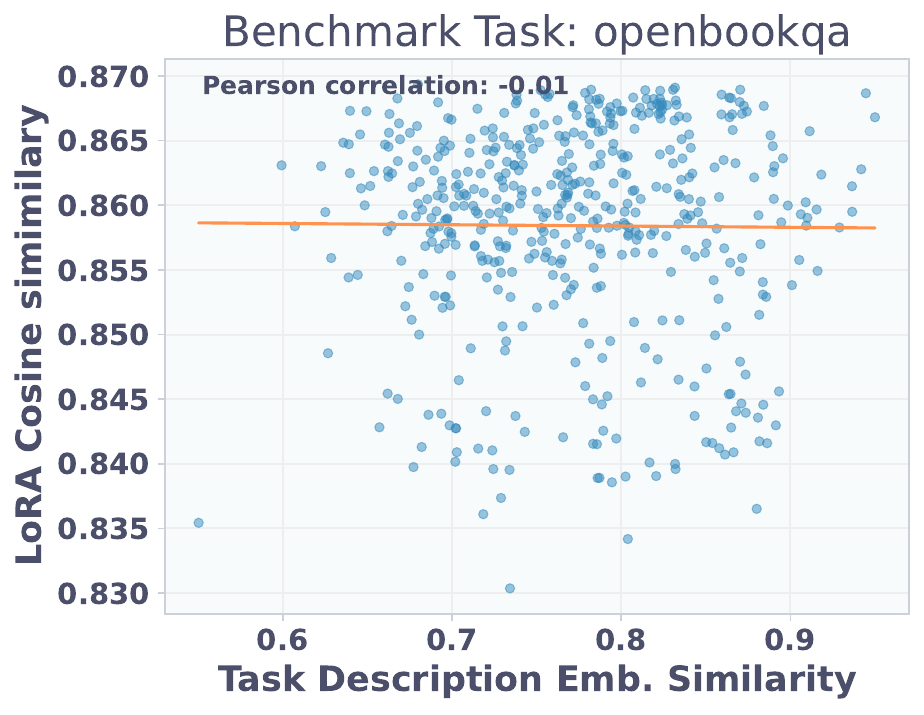}
    \\
    {
    \includegraphics[width=0.31\linewidth]{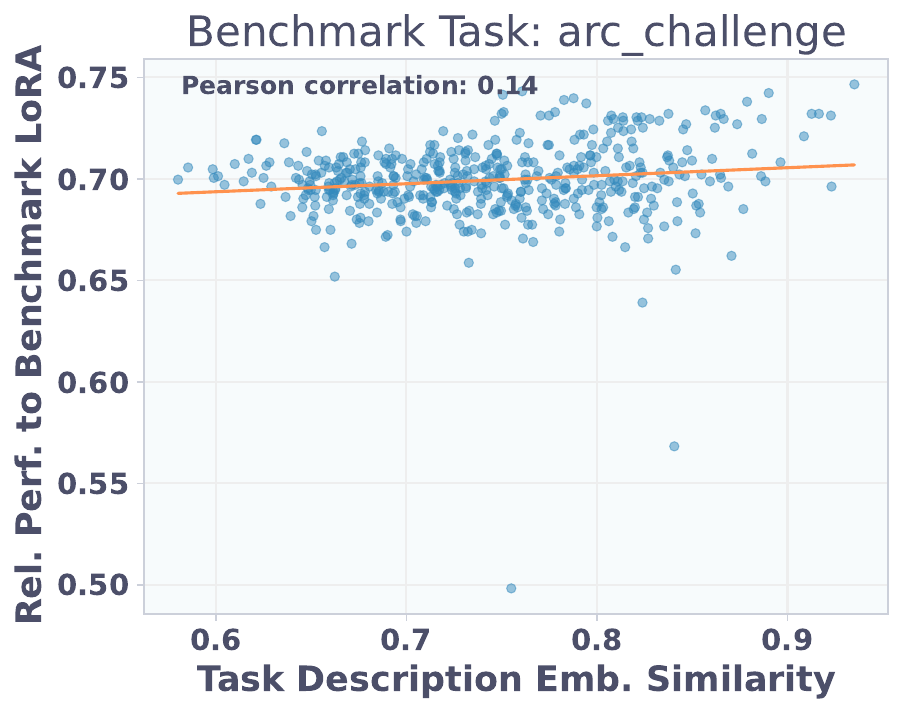}
    }
    {
    \includegraphics[width=0.31\linewidth]{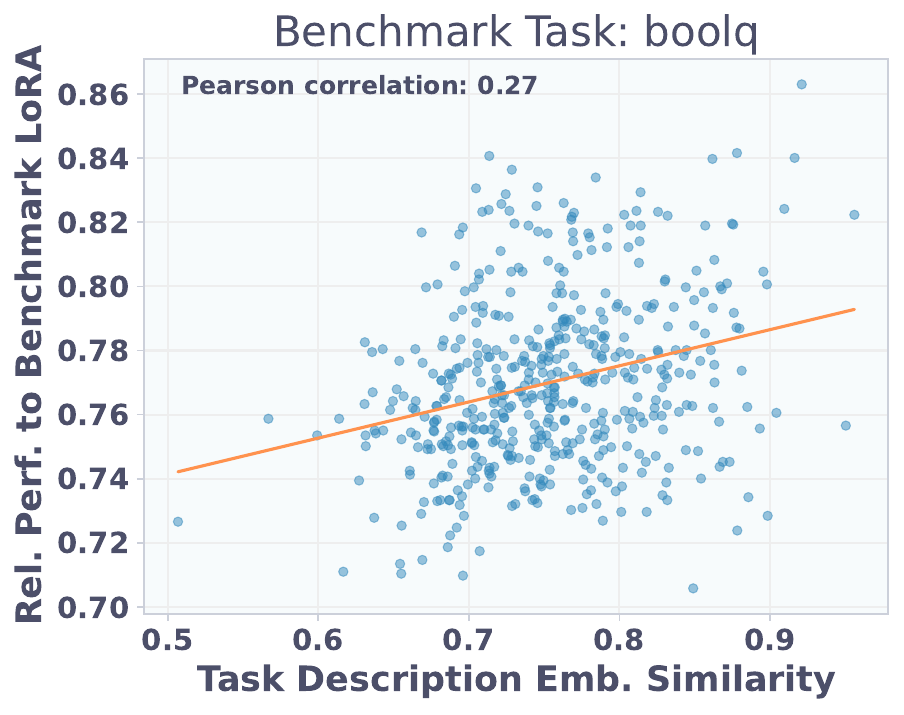}
    }
    {
    \includegraphics[width=0.31\linewidth]{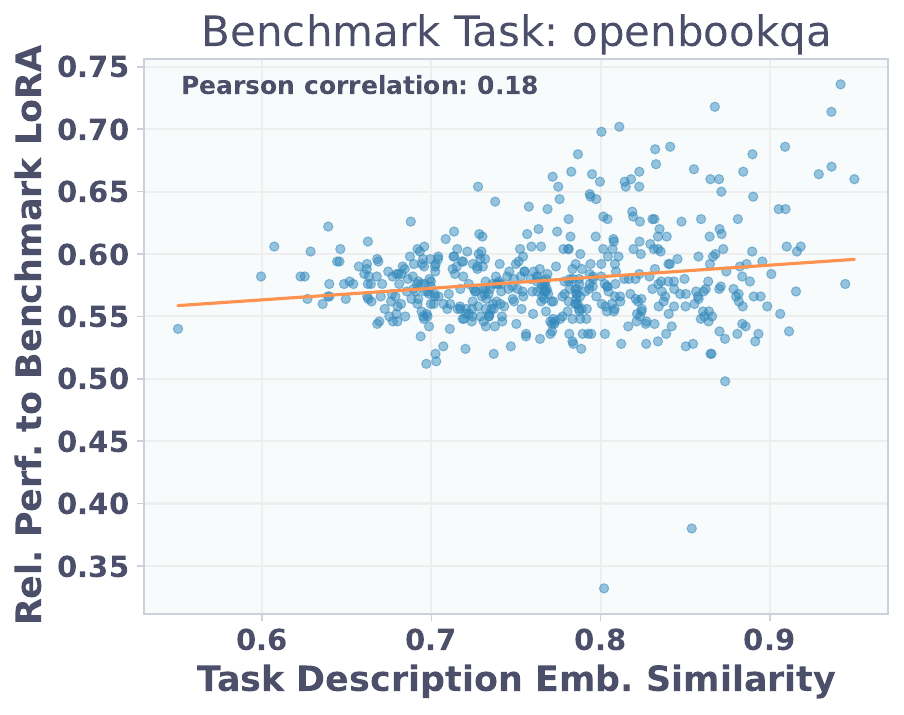}
    }
    \vspace{-0.2cm}
    \caption{\textbf{Top row:} Each plot shows the similarity between a benchmark \lora{} adapter and 479 SNI-trained adapters in the weight space (y-axis) against their similarity in the task embedding space (x-axis). \textbf{Bottom row:} Each plot shows SNI-trained adapters' performance relative to a benchmark adapter (y-axis) with the same x-axis. We can see that \lora{}s with similar description embeddings to the benchmarks' perform better in those benchmarks, suggesting their shared functionalities. However, \lora{}s with similar functionalities are not nearby in the parameter space.}
    \label{fig:ab_lora_corr}
    \vspace{-0.5cm}
\end{figure}
Here, we investigate the relationship between \lora{} adapters by inspecting their similarity in the parameter space, performance on the benchmarks, and similarity of their description embeddings. To measure adapter similarity, we compute the cosine similarity of the concatenation of flattened low-rank $A$ and $B$ matrices of all layers.
In the top row of \cref{fig:ab_lora_corr}, we plot the adapters' similarity against task description similarity (using the mean embedding of each task).
We find no correlation between the cosine similarity of the adapters' weights (y-axis) and the task embedding similarity (x-axis) indicated by near-zero Pearson correlation coefficients.

In the bottom row of \cref{fig:ab_lora_corr}, we change the y-axis from adapters' relative benchmark performance to benchmark-specific adapters. We find a positive correlation between the relative benchmark performance of SNI-trained adapters and the task embedding similarity. 
That is, adapters perform better on a benchmark if their task descriptions are similar to those of the benchmark. However, despite their similar functionalities, adapters with similar descriptions are not similar in the parameter space. We believe that this relationship has a significant impact on the limited generalization of reconstruction-trained \ours{}. We further discuss this topic in \cref{app:full_matrix_vs_ab}.

\section{Hyperparameter Settings}\label{app:hyper_params}
\begin{table}[ht]
    \centering
    \caption{Hyperparameters for training a task-specific \lora{} adapter.}
    \label{tab:hyperparams}
    \scalebox{0.9}{
    \begin{tabular}{lccc}
    \toprule
    \textbf{Hyperparameters} & \textbf{Task-specific LoRA} & \textbf{\ours{} (SFT)} & \textbf{\ours{} (recon)} \\ \midrule
    Batch size                  & $8$                & $8$       & Number of the target \lora{}s       \\
    Gradient accumulation steps & $1$                & $1$      & $1$      \\
    Max learning rate           & $8 \times 10^{-5}$ & $2.5\times10^{-5}$ & $10^{-3}$ \\
    Max gradient norm           & $1.0$ & $1.0$ & $1.0$    \\
    NEFTune noise alpha         & $5.0$ & $5.0$ & $5.0$    \\
    Warmup fraction             & $0.1$ & $0.1$ & $0.1$    \\
    Learning rate scheduler  & Linear with warm up         & Linear with warm up      & Linear with warm up      \\ \bottomrule   
    \end{tabular}%
    }
\end{table}
\begin{listing}[ht]
\begin{minted}
[
frame=lines,
framesep=2mm,
]
{json}
{
  "alpha_pattern": {},
  "auto_mapping": null,
  "base_model_name_or_path": "models/Mistral-7B-Instruct-v0.2",
  "bias": "none",
  "fan_in_fan_out": false,
  "inference_mode": true,
  "init_lora_weights": true,
  "layer_replication": null,
  "layers_pattern": null,
  "layers_to_transform": null,
  "loftq_config": {},
  "lora_alpha": 16,
  "lora_dropout": 0.05,
  "megatron_config": null,
  "megatron_core": "megatron.core",
  "modules_to_save": null,
  "peft_type": "LORA",
  "r": 8,
  "rank_pattern": {},
  "revision": null,
  "target_modules": [
    "q_proj",
    "v_proj"
  ],
  "task_type": "CAUSAL_LM",
  "use_dora": false,
  "use_rslora": true
}
\end{minted}
\caption{The parameter-efficient fine-tuning (PEFT) config for all \lora{} adapters.}
\label{code:peft_config}
\end{listing}
\cref{tab:hyperparams,code:peft_config} show the training configuration of all models trained in this work. For \lora{} reconstruction training, each prediction target is an entirety of a \lora{} adapter. That is, there is a total of 479 training samples for 479 SNI tasks. Thus, we increase the epochs to $100,000$ to ensure that \ours{} converges.

\newpage
\section{Additional Details of \ours{} Architectures}\label{app:architecture}
\begin{listing}[ht]
\begin{minted}
[
frame=lines,
framesep=2mm,
fontsize={\fontsize{8}{8}\selectfont},
]
{python}
Hypermod: HyperModulator(
  (task_encoder): TaskEncoder(
    (mlp): Sequential(
      (0): Linear(in_features=1024, out_features=64, bias=True)
      (1): LayerNorm((64,), eps=1e-05, elementwise_affine=True)
    )
  )
  (layer_depth_encoder): Sequential(
    (0): Embedding(32, 32)
    (1): LayerNorm((32,), eps=1e-05, elementwise_affine=True)
  )
  (layer_type_encoder): Sequential(
    (0): Embedding(2, 32)
    (1): LayerNorm((32,), eps=1e-05, elementwise_affine=True)
  )
  (mixer): Sequential(
    (0): Linear(in_features=128, out_features=512, bias=True)
    (1): SiLU()
    (2): Dropout(p=0.05, inplace=False)
    (3): Linear(in_features=512, out_features=128, bias=True)
    (4): SiLU()
    (5): Dropout(p=0.05, inplace=False)
  )
  (mlp1): MLPResidualBlock(
    (mlp): Sequential(
      (0): LayerNorm((128,), eps=1e-05, elementwise_affine=True)
      (1): Linear(in_features=128, out_features=512, bias=True)
      (2): SiLU()
      (3): Dropout(p=0.05, inplace=False)
      (4): Linear(in_features=512, out_features=128, bias=True)
      (5): SiLU()
      (6): Dropout(p=0.05, inplace=False)
    )
  )
  (mlp2): MLPResidualBlock(
    (mlp): Sequential(
      (0): LayerNorm((128,), eps=1e-05, elementwise_affine=True)
      (1): Linear(in_features=128, out_features=512, bias=True)
      (2): SiLU()
      (3): Dropout(p=0.05, inplace=False)
      (4): Linear(in_features=512, out_features=128, bias=True)
      (5): SiLU()
      (6): Dropout(p=0.05, inplace=False)
    )
  )
  (mlp3): Sequential(
    (0): LayerNorm((128,), eps=1e-05, elementwise_affine=True)
    (1): Linear(in_features=128, out_features=512, bias=True)
    (2): SiLU()
    (3): Dropout(p=0.05, inplace=False)
    (4): Linear(in_features=512, out_features=512, bias=True)
    (5): SiLU()
  )
)
\end{minted}
\caption{Detailed backbone architecture.}
\label{code:backbone}
\end{listing}
\begin{listing}[ht]
\begin{minted}
[
frame=lines,
framesep=2mm,
fontsize={\fontsize{8}{8}\selectfont},
]
{python}
(AB_emb): ParameterDict(
      (q_proj): Object of type: ParameterDict
      (v_proj): Object of type: ParameterDict
    (q_proj): ParameterDict(
        (A): Parameter containing: [torch.cuda.FloatTensor of size 128]
        (B): Parameter containing: [torch.cuda.FloatTensor of size 128]
    )
    (v_proj): ParameterDict(
        (A): Parameter containing: [torch.cuda.FloatTensor of size 128]
        (B): Parameter containing: [torch.cuda.FloatTensor of size 128]
    )
  )

(rank_emb): Sequential(
    (0): Embedding(8, 128)
    (1): LayerNorm((128,), eps=1e-05, elementwise_affine=True)
  )
\end{minted}
\caption{Detailed A/B and rank embedding of \ours{}.}
\label{code:ab_emb_and_rank_emb}
\end{listing}
\cref{code:backbone,code:ab_emb_and_rank_emb} show the details of the backbone of \ours{}. Specifically, the size of the module and layer embedding ($E[m]$ and $E[l]$) is 32D. Together, they form a dictionary of 34 learnable embeddings (32 layers + 2 target modules). The task encoder is a linear layer that takes in a text embedding (1024D for the \texttt{gte} embedding and 4096D for \texttt{Mistral} embedding) and outputs a 64D vector. The encoded task, module, and layer embedding are concatenated and then fed into \texttt{mlp0} followed by a residual MLP block \texttt{mlp1}. At this point, for \March{} and \Sarch{}, we add a 128D A/B embedding to the residual stream. The output is then fed to another residual MLP block \texttt{mlp2}. At this point, for \Sarch{}, we add a 128D rank embedding to the residual stream. After this, we feed the activation to the last MLP block. The output of the last MLP block is then fed to a linear head, whose output size is as follows:
\begin{itemize}
    \item \Larch{}: $2 \times r \times d$ giving both $A$ and $B$ matrices
    \item \March{}: $r \times d$ giving a low-rank matrix $A$ or $B$ depending on the A/B embedding
    \item \Sarch{}: $d$ giving a rank of a low-rank matrix depending on both the A/B embedding and the rank embedding.
\end{itemize}
For ease of explanation, we assume $d$ is the same for the input and the output space of a linear transformation. In practice, $d_\text{in} = d_\text{out} = 4096$ for \texttt{q\_proj} module and $d_\text{in} = 4096, d_\text{out}=1024$ for \texttt{v\_proj} module. $r=8$ for all adapters in this work.
Finally, we list the number of trainable parameters of each architecture: $55,252,992$ for \Larch{}, $34,282,240$ for \March{}, $4,923,392$ for \Sarch{}, $3,407,872$ for \lora{}.

\section{\ours{} Intialization}\label{app:hyper_init}
We use \emph{Bias-HyperInit} \citep{beck2023bias-hyperinit} to initialize \Larch{} \ours{}. Bias-HyperInit initializes the linear output head of the hypernetwork such that the weights are all zeros and the bias matches the initialization of the underlying layers. In our work, this corresponds to the output bias of the \Larch{} hypernetwork being initialized to $U(-\frac{1}{d}, \frac{1}{d})$ for the $A$ head and all zero for the $B$ head to match the initialization of traditional \lora{}. 
For other architectures, we aim to match the gradient magnitude to \Larch{} at the beginning of training. That is, for \March{} architecture, we initialize the bias of the output head to $U(-\frac{1}{\sqrt{2}d}, \frac{1}{\sqrt{2}d})$. Finally, \Sarch{} output bias is initialized to $U(-\frac{1}{\sqrt{r2}d}, \frac{1}{\sqrt{r2}d})$. Without this explicit hypernetwork initialization, the training is unstable, and often leads to failed training runs.

\section{Training Details}\label{app:reproducibility}
All models trained in this work fit in a single H100 GPU (80GB of VRAM). Notably, SFT requires much more memory because of the need to backpropagate the gradient through the base LLM. Reconstruction training, on the other hand, should be possible in a modern consumer-grade GPU.

For reconstruction training, we fix the training epochs to be 100K but scale the batch size to match the number of target \lora{} adapters. This means the model trains much faster for a lower number of target \lora{}s while maintaining the same number of optimizer steps. For reference, training to reconstruct 9 benchmark-specific \lora{}s takes around 10 minutes to complete, while training to reconstruct 479 SNI \lora{} adapters takes around 10 hours.

For SFT training with fixed compute budget, we aim to keep the number of optimizer steps the same as we do for reconstruction training. However, since we cannot fit all fine-tuning samples, we scale the number of epochs inverse to the number of training tasks.

Additionally, for reconstruction training, instead of predicting the weights directly, \ours{} learns to predict the z-score of a normal distribution of each weight entry in the low-rank $\textbf{A}, \textbf{B}$ matrices. At test time, the output is multiplied by the standard deviation of each element before adding to the mean, converting the prediction to the correct scale.

\section{Ad-hoc FLOPs Analysis}\label{app:flops_analysis}
Let $S$ be the sequence length, $H$ be the hidden size, and $L$ be the number of layers of a Transformer-based LLM.
We use the following equations for computing the matrix multiplications (GEMMs) FLOPs \citep{korthikanti2023reducing}.

\textbf{FLOPs for Self-Attention (per layer):} $8 \times S \times H ^ 2 + 4 \times H \times S ^ 2$

\textbf{FLOPs for FFN (per layer):} $16 \times S \times H ^ 2$

\textbf{Per Transformer Block Total FLOPs:} $24 \times S \times H ^ 2 + 4 \times H \times S ^ 2$

\textbf{Setup for comparison:}
\vspace{-0.5cm}
\begin{itemize}
    \setlength{\itemsep}{-8pt}
    \item 3-shot ICL examples are approximately 256 tokens long
    \item Question instances are approximately 64 tokens long
    \item Task descriptions are approximately 48 tokens long
    \item We consider one question instance as the main input to the base model
    \item We only consider input tokens for the FLOPs calculation
    \item We use `Mistral-7B-Instruct-v0.2` as the base model (S = 256 + 64 (3-shot ICL + question instance), H = 4096, L = 32)
    \item When the based model is used with T2L, we do not include 3-shot ICL (S = 64 (question instance), H = 4096, L = 32)
    \item We use `gte-large-en-v1.5` as the task description encoder (S = 48 (task description), H = 1024, L = 24)
    \item We use the M hypernetwork architecture detailed in the Appendix F
\end{itemize}
\subsection{T2L per instance FLOPs}

\textbf{gte-large-en-v1.5:} FLOPs $= 24 \times (24 \times 48 \times 1024 ^ 2 + 4 \times 1024 \times 48 ^ 2) = 0.029$ TFLOPs/instance

\textbf{Hypernetwork (M):} FLOPs $= 2 \times 1024 \times 64 + 4 \times 4 \times 128 \times 512 + 128 \times 4096 \times 8 = 0.000005$ TFLOPs/instance

\textbf{Base LLM w/o ICL:} FLOPs $= 32 \times (24 \times 64 \times 4096 ^ 2 + 4 \times 4096 \times 64 ^ 2) = 0.827$ TFLOPs/instance

\textbf{Total FLOPs} $= 0.029 + 0.000005 + 0.827 = \mathbf{0.856005}$ \textbf{TFLOPs/instance}

\subsection{Base LLM with 3-shot ICL}
\textbf{Total FLOPs} = $32 \times (24 \times (256 + 64) \times 4096 ^ 2 + 4 \times (4096) \times (256 + 64) ^ 2) = \mathbf{4.177}$ \textbf{TFLOPs/instance}

Based on this calculation, we can see that the adaptation cost of \ours{} is significantly cheaper than 3-shot ICL—more than 4x FLOPs reduction, saving compute within the first question instance.

\section{Training and Evaluation Datasets}\label{app:datasets}
We use 500 SNI datasets publicly available at \url{https://huggingface.co/Lots-of-LoRAs}. 479 tasks are used for training and the rest for evaluation. Specifically, we use the following evaluation tasks: task\_035, task\_039, task\_1557, task\_202, task\_304, task\_362, task\_614, task\_701, task\_706, task\_710, task\_726.
For the in-context learning baseline, we use 3-shot in-context examples taken from the training split of each benchmark except MBPP that has an explicit split for in-context prompting. HumanEval only has the test split, therefore it is always evaluated against in the zero-shot manner.
\begin{figure}[H]
    \centering
    \includegraphics[page=9,width=0.8\linewidth]{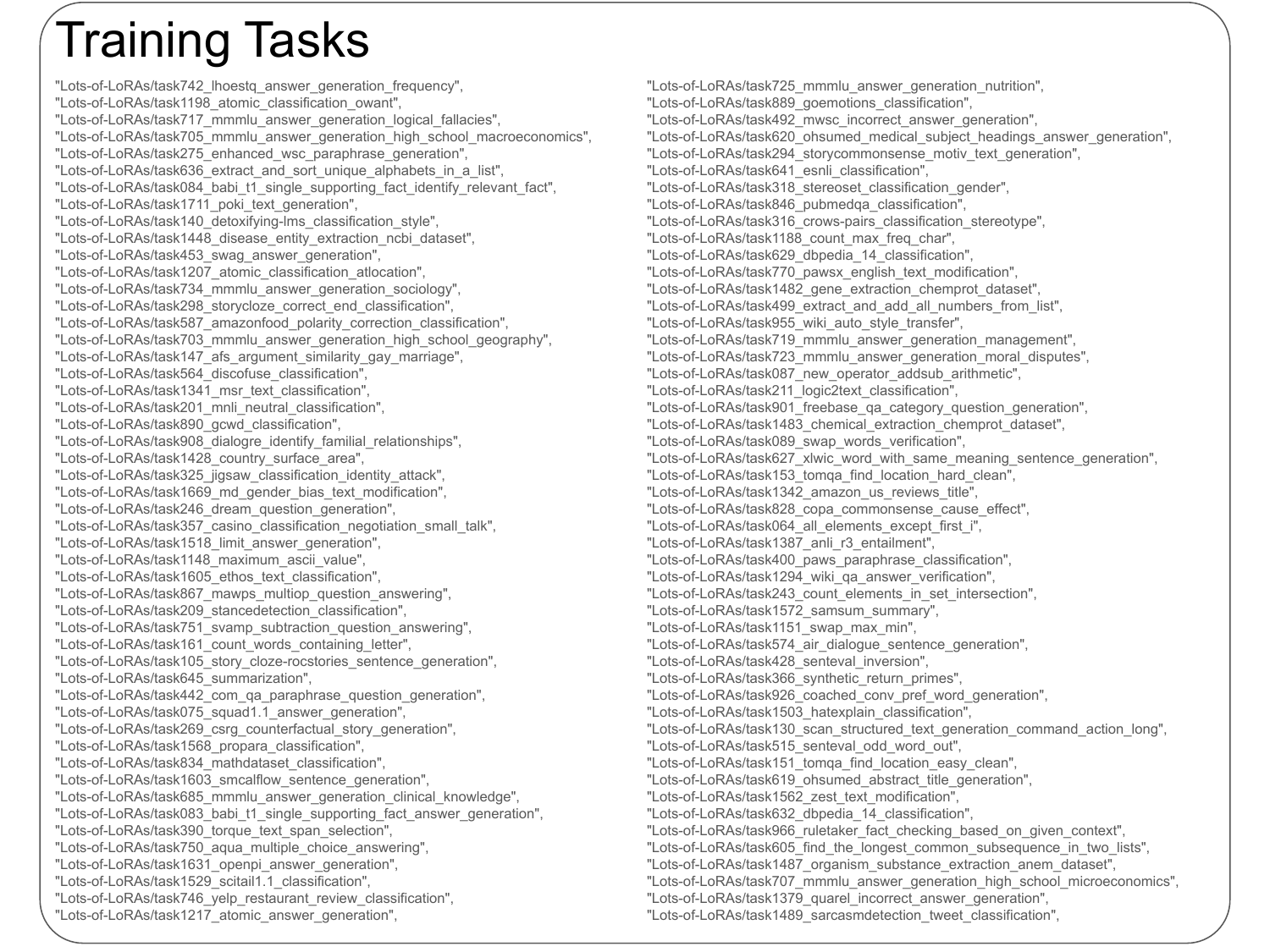}
    \\
    \includegraphics[page=10,width=0.8\linewidth]{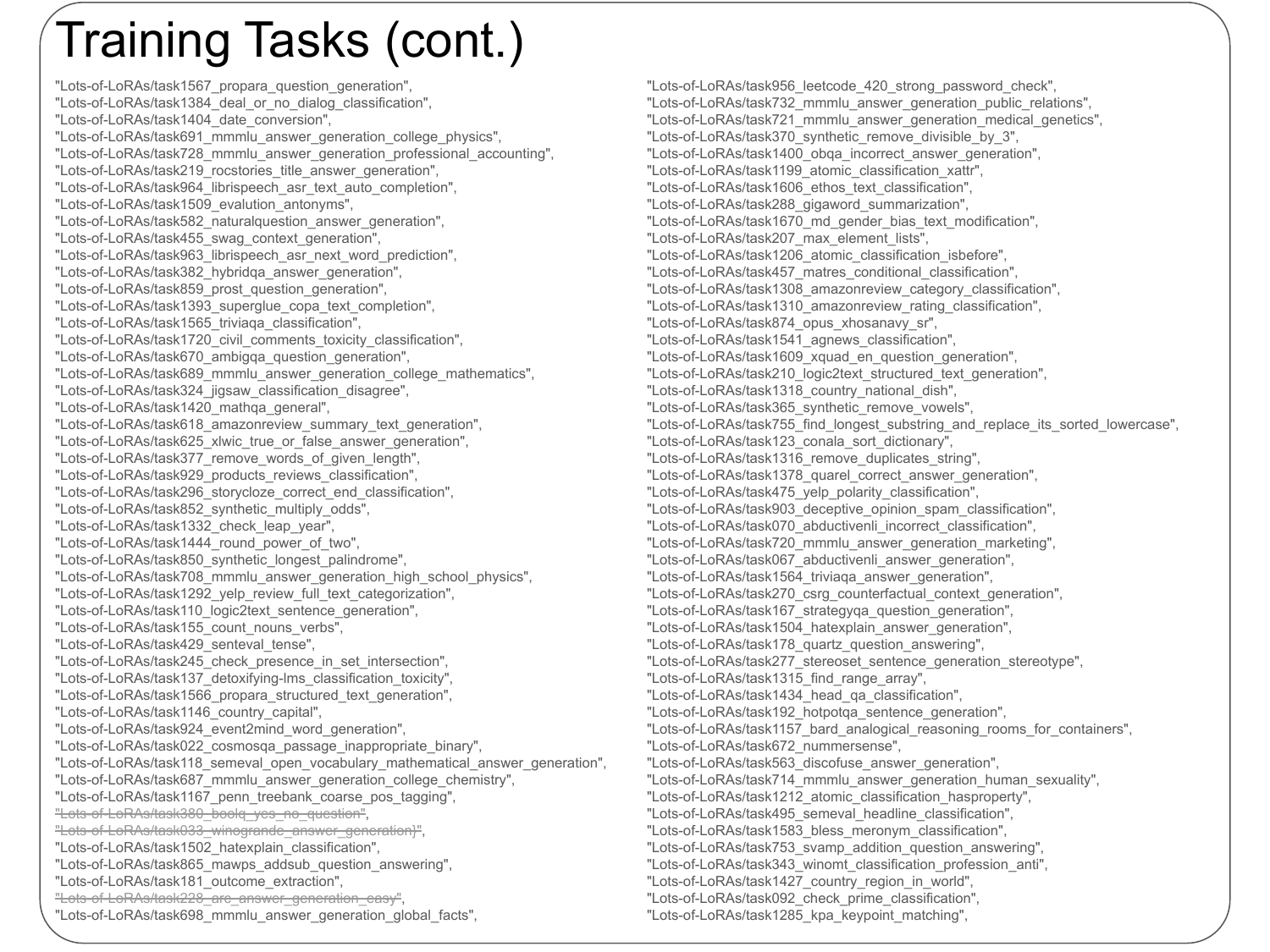}
    \caption{Training tasks from Lots-of-LoRAs (based on the SNI dataset) used for training the Text-to-LoRA model. The struck out names indicate removed tasks due to benchmark contamination.}
\end{figure}
\begin{figure}[H]
    \centering
    \includegraphics[page=11,width=0.8\linewidth]{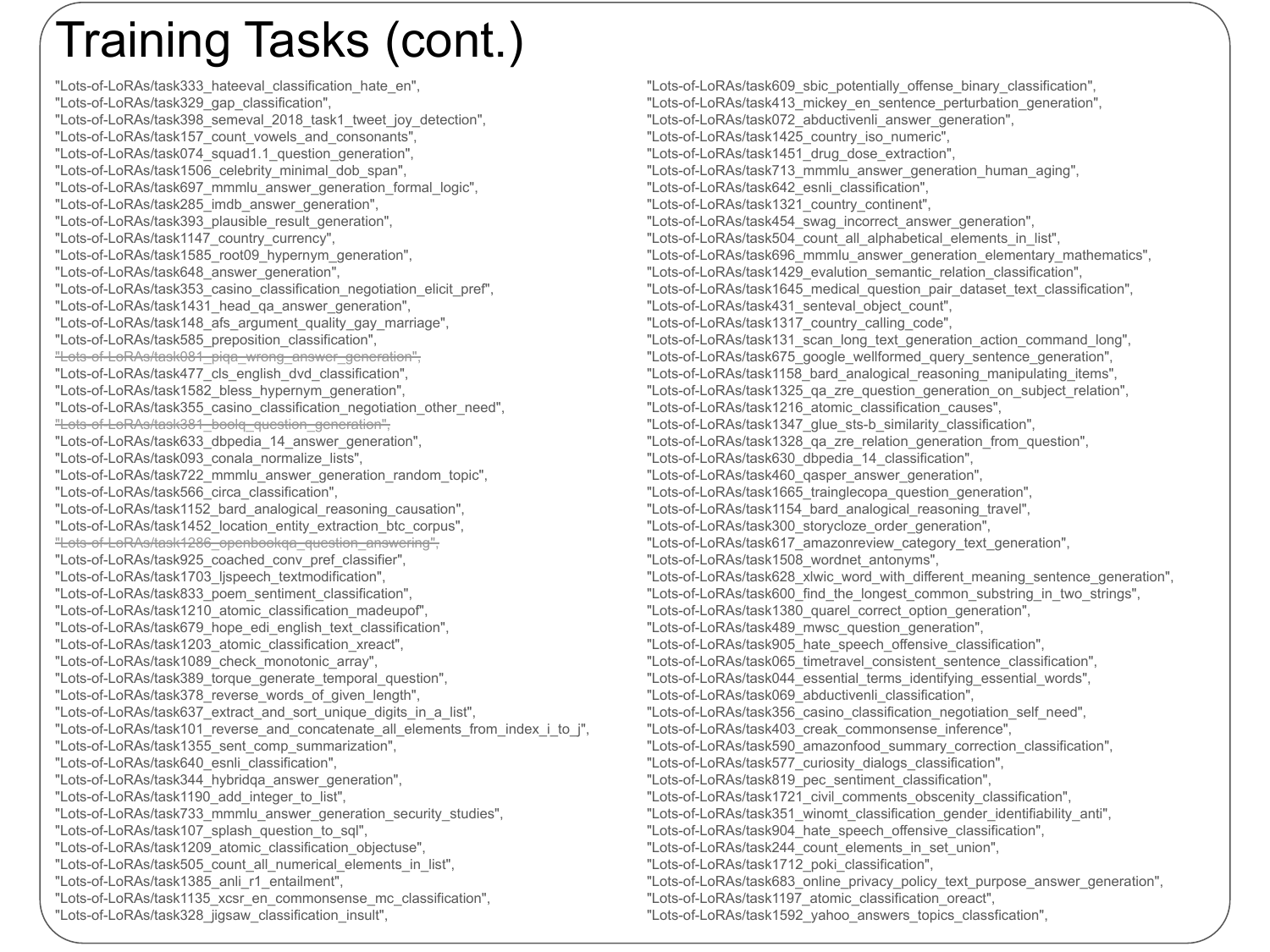}
    \\
    \includegraphics[page=12,width=0.8\linewidth]{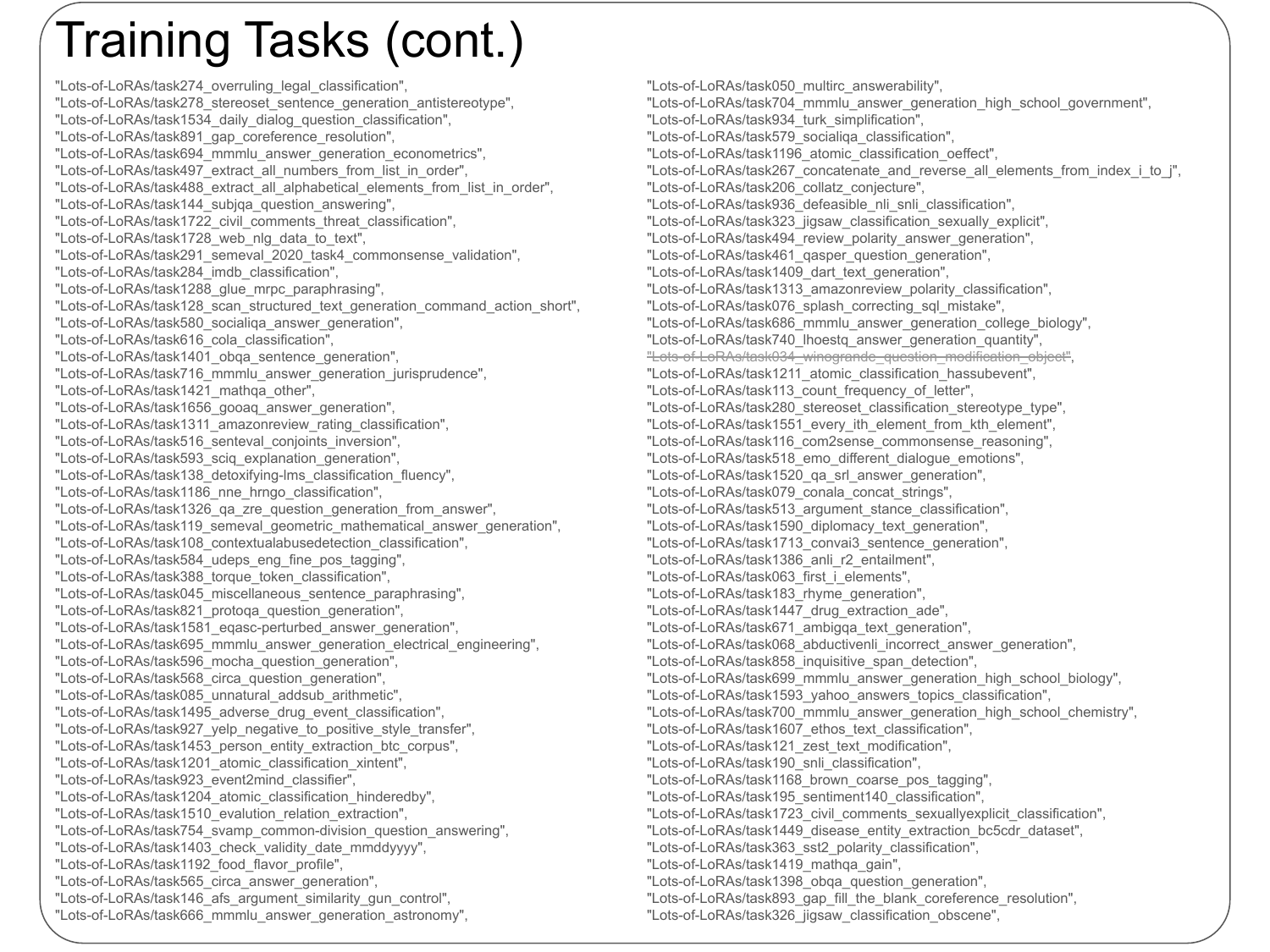}
    \caption{Training tasks from Lots-of-LoRAs (based on the SNI dataset) used for training the Text-to-LoRA model. The stricken out names indicate removed tasks due to benchmark contamination.}
\end{figure}
\begin{figure}[H]
    \centering
    \includegraphics[page=13,width=0.8\linewidth]{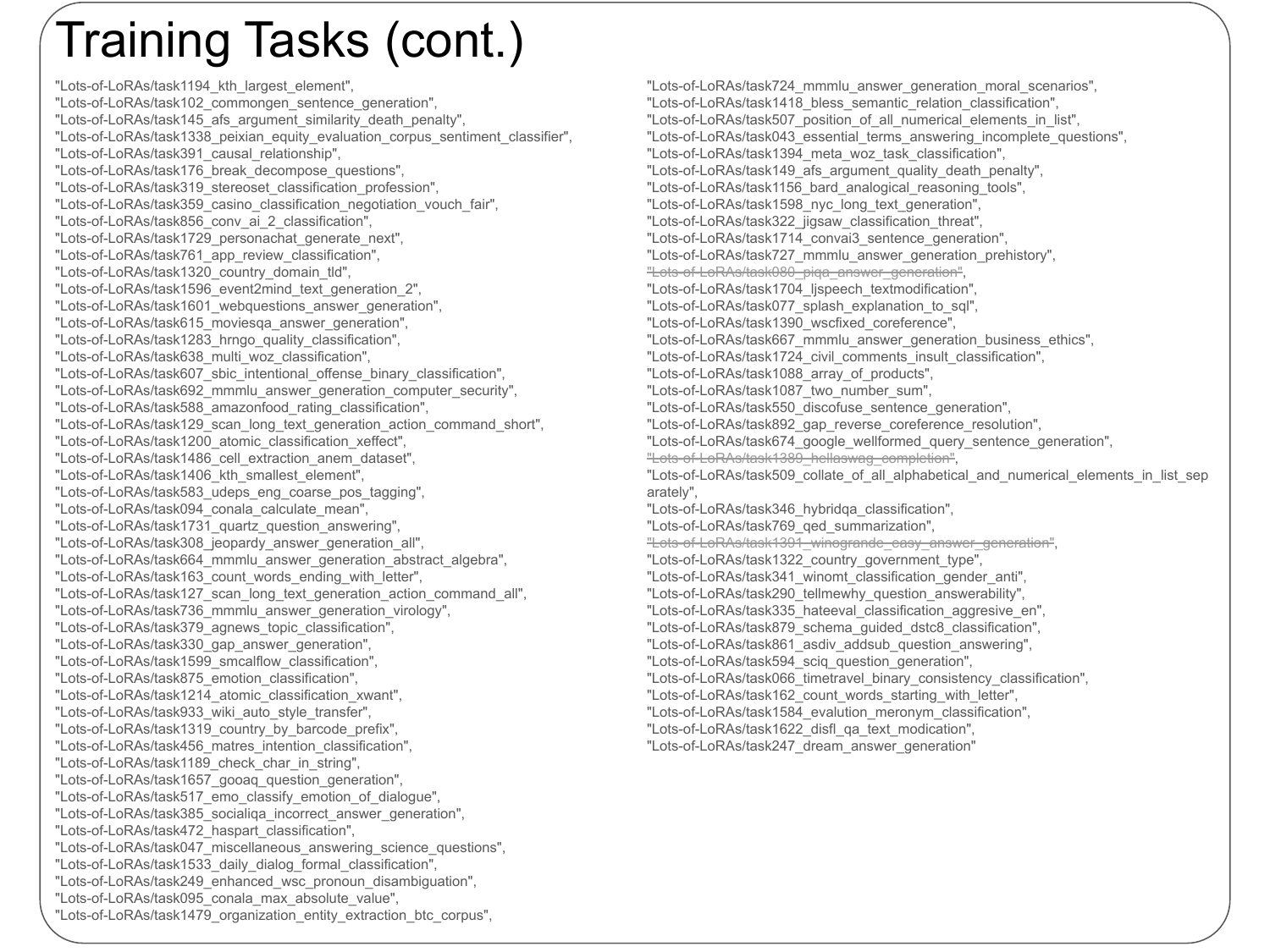}
    \caption{Training tasks from Lots-of-LoRAs (based on the SNI dataset) used for training the Text-to-LoRA model. The stricken out names indicate removed tasks due to benchmark contamination.}
\end{figure}
\begin{figure}[H]
    \centering
    \includegraphics[page=14,trim=0cm 11cm 0cm 0cm,clip,width=0.8\linewidth]{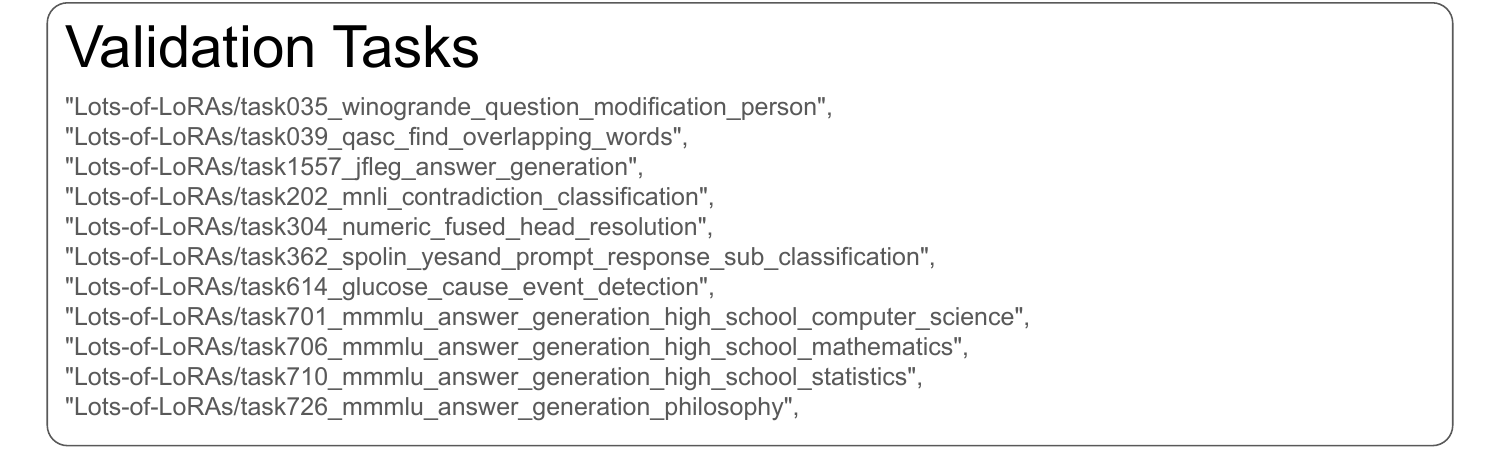}
    \caption{Validation tasks used during the training of the Text-to-LoRA model.}
\end{figure}
\subsection{Benchmark Details}
Every benchmark used in the experiments is publicly available in HuggingFace dataset space. We evaluate the models on the benchmarks detailed as follows.

\subsubsection{GSM8K}
We evaluate the models on the test split, using chain-of-thought response pre-filling: \textit{"Let's think step by step."}

\subsubsection{HumanEval and MBPP}
We use the \texttt{evalplus} library \citep{evalplus} for coding evaluation. For both MBPP and HumanEval, we use the following response pre-fill: \texttt{```python}

\subsection{Question-Answering Tasks}
The rest of the benchmarks are question-answering based tasks. In these tasks, we do not use response-prefilling. Instead, each task has a specific instruction template shown in \cref{code:instruction_templates}.
\begin{listing}[ht]
\begin{minted}
[
frame=lines,
framesep=2mm,
fontsize={\fontsize{8}{8}\selectfont},
]
{python}
OQA_TEMPLATE = (
    "Complete the following passage or answer the question by choosing the correct choice.\n\n"
    "{question_stem}\n\n"
    "{choices[label][0]}: {choices[text][0]}\n{choices[label][1]}: {choices[text][1]}\n"
    "{choices[label][2]}: {choices[text][2]}\n{choices[label][3]}: {choices[text][3]}\n\n"
    "You must respond with the letter corresponding to the correct choice (A,B,C,D)"
    " without any explanation."
)
ARC_TEMPLATE = (
    "Answer the question below by choosing the correct choice.\n\n"
    "{question}\n\n"
    "{choices[label][0]}: {choices[text][0]}\n{choices[label][1]}: {choices[text][1]}\n"
    "{choices[label][2]}: {choices[text][2]}\n{choices[label][3]}: {choices[text][3]}\n\n"
    "You must respond with the letter corresponding to the correct choice without any explanation."
)
HSWAG_TEMPLATE = (
    "You are provided with an incomplete passage below as well as 4 choices of continuation "
    "with only one of them being the correct ending. "
    "Treat the endings as being labelled 0, 1, 2, 3 in order.\n\n"
    "Passage: {ctx}\n\n"
    "0: {endings[0]}\n"
    "1: {endings[1]}\n"
    "2: {endings[2]}\n"
    "3: {endings[3]}\n\n"
    "You must respond with the only number corresponding to the correct ending (0,1,2,3)"
    " for the passage without any explanation."
)
PIQA_TEMPLATE = (
    "Choose the option that either answers the question, completes the sentence,"
    " or solves the problem. "
    "Pay attention to the properties of the objects in the question and how they interact with "
    "each other. "
    'If both options are correct, choose the one that is more convenient or more common.
    "\n\n"""{goal}"""\n\n0: {sol1}\n1: {sol2}\n\n"
    "You must respond with either `0` or `1` without any explanation."
)
WINOGRANDE_TEMPLATE = (
    "Given the following situation:\n\n{sentence}\n\nWhich option is correct?\n\n"
    "Option 1: {option1}\n\nOption 2: {option2}\n\n"
    "You must respond with either `1` or `2` without any explanation."
)
BOOLQ_TEMPLATE = (
    "{passage}\n\nQuestion: {question}?\n\nPlease answer with either `true` or `false` "
    "without any explanation."
)
\end{minted}
\caption{Instruction templates of QA-based benchmark tasks.}
\label{code:instruction_templates}
\end{listing}

\section{Utilizing Full Adaptation Matrix vs Low-Rank Matrices}\label{app:full_matrix_vs_ab}
\begin{figure}[ht]
    \centering
    \includegraphics[width=0.325\linewidth]{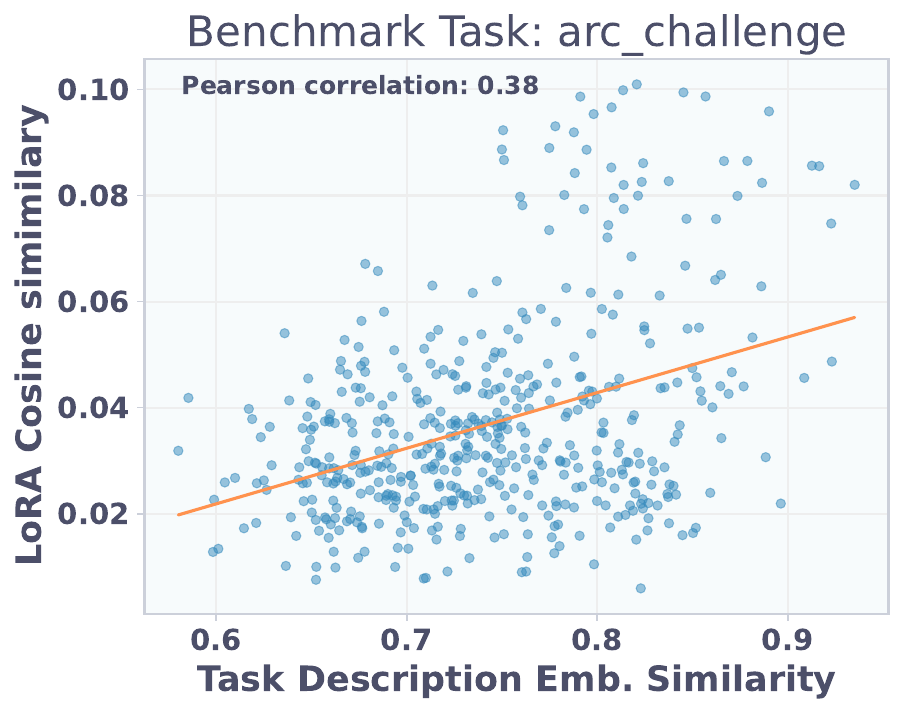}
    \includegraphics[width=0.325\linewidth]{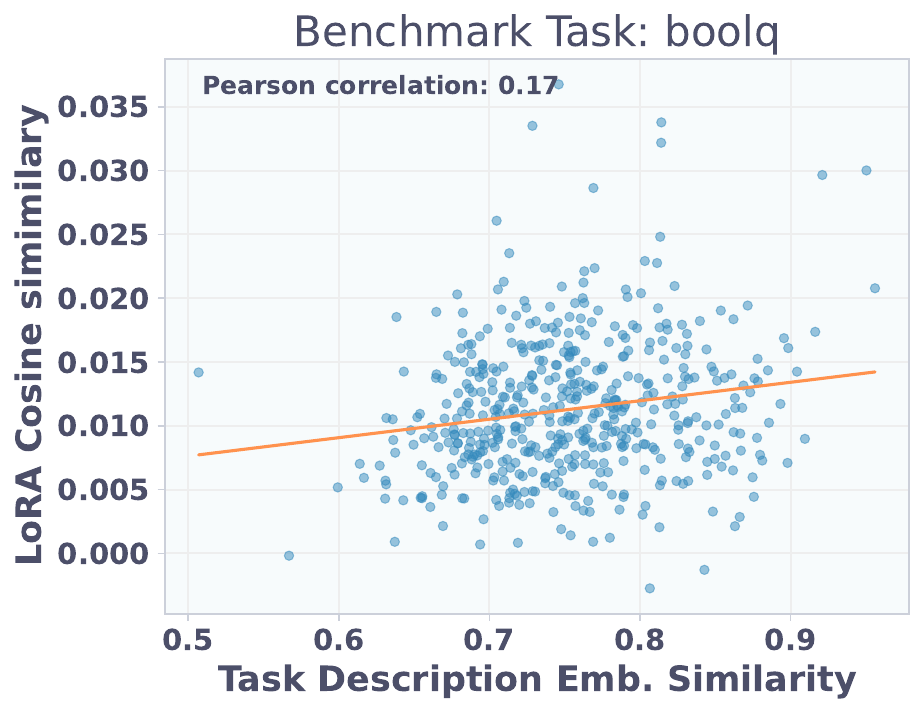}
    \includegraphics[width=0.325\linewidth]{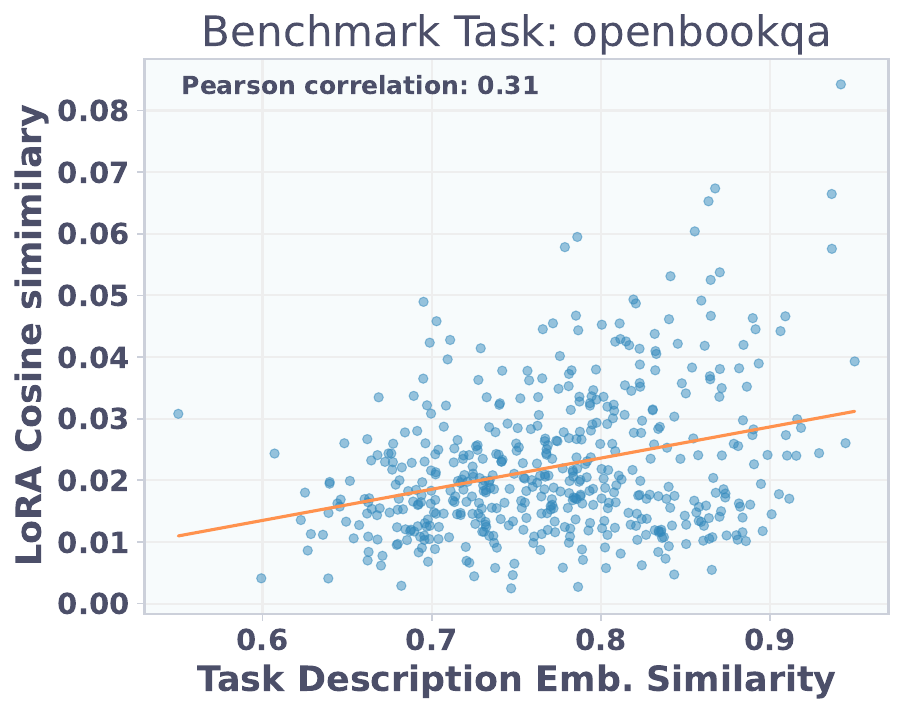}
    \caption{Each plot shows the similarity between a benchmark \lora{} adapter and 479 SNI-trained adapters in the $\Delta W$ weight space. There is a positive correlation between the two variables indicated by small positive Pearson correlation coefficients.}
    \label{fig:full_matrix_lora_corr}
\end{figure}
Similar to \cref{fig:ab_lora_corr}, \cref{fig:full_matrix_lora_corr} shows the similarity of SNI adapters to benchmark-specific adapters, but instead of using the concatenation of flattened $A$ and $B$ matrices, we use flattened $\Delta W$ instead. With the change, we find a positive correlation between the task embedding similarity and the adapter similarity in the weight space. This is likely because, for a given $\Delta W$ matrix, there are many possible permutations of low-rank matrices $A$ and $B$. This suggests that if we compute the reconstruction loss in the full adaptation matrix space, reconstruction-trained \ours{} could generalize better. However, we empirically find that it does not outperform \ours{} trained to reconstruct low-rank matrices at zero-shot \lora{} generation.

\section{Generating Task Descriptions with a Foundation Language Model}\label{app:gpt_prompt}
\begin{figure}[H]
    \centering
    \includegraphics[page=1,trim={1cm 2cm 1cm 0cm},clip,width=0.8\linewidth]{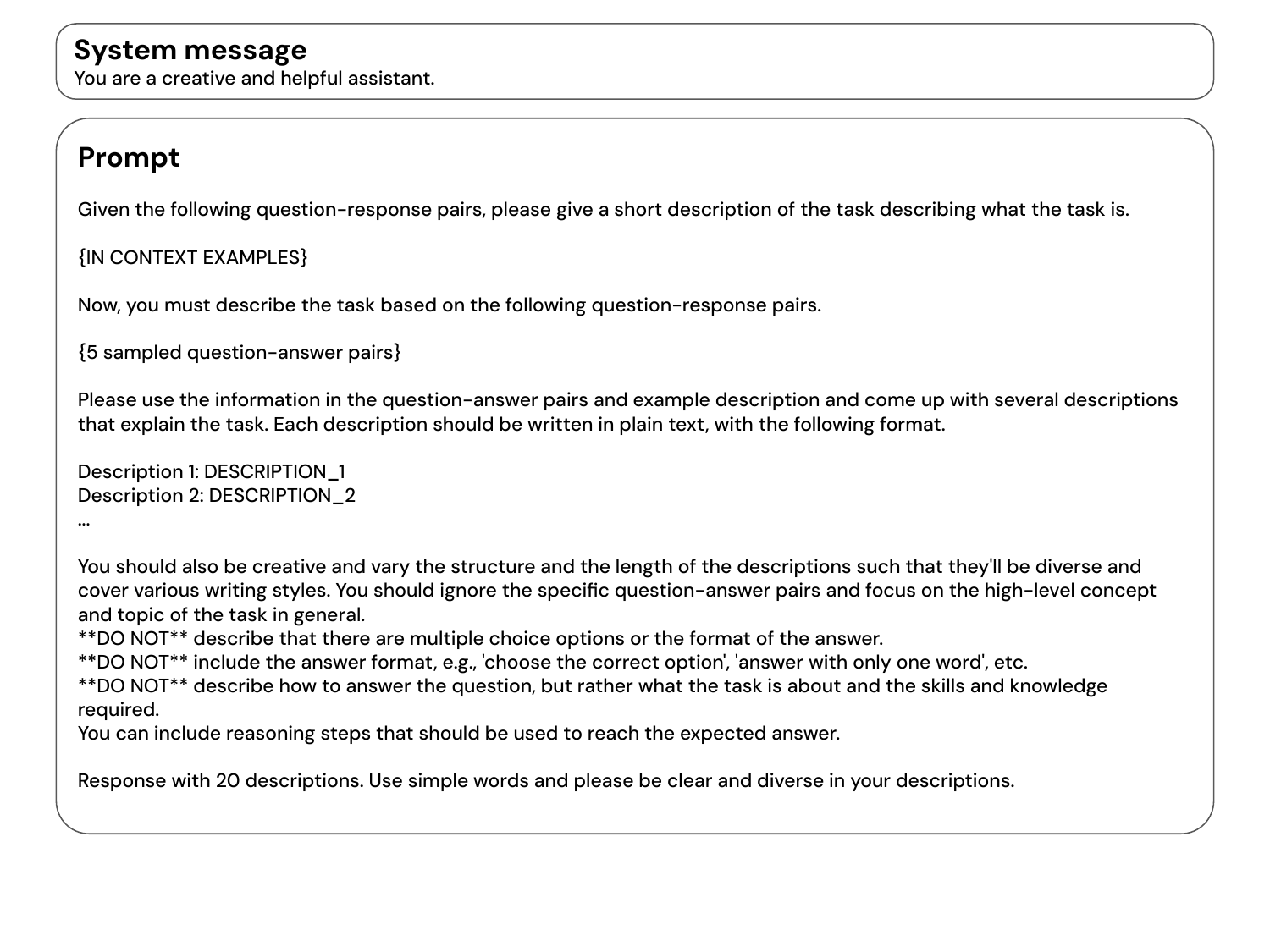}
    \includegraphics[page=2,trim={1cm 1cm 1cm 1cm},clip,width=0.8\linewidth]{figs/gpt_prompt.pdf}
    \caption{The prompt template used to query \texttt{GPT-4o mini} for task descriptions.}
    \label{fig:gpt_prompt}
\end{figure}

We automate task description generation for each task by leveraging powerful closed-source language models~\citep{achiam2023gpt}. We query \texttt{GPT-4o mini} with carefully constructed prompts that incentivize diversity to facilitate downstream generalization.
In particular, we generate 200 descriptions per task by querying the model 10 times, each time asking for 20 descriptions given randomly sampled five question-answer pairs from the task.
We leverage in-context learning by providing examples of question-answer pairs with matching descriptions.
Finally, we also designed our prompts to avoid overly verbose responses and unnecessary information, such as explicit mentions of answer formats and additional instructions. We use the generated descriptions for the training and benchmark tasks. \cref{fig:gpt_prompt} shows the exact prompt used for querying \texttt{GPT-4o mini} for task descriptions.

\newpage
\section{Example of Task Descriptions}
Here, we provide examples of task descriptions used in the experiments.
\begin{figure}[H]
    \centering
    \includegraphics[page=3,trim={0cm 1cm 0cm 1cm},clip,width=1.0\linewidth]{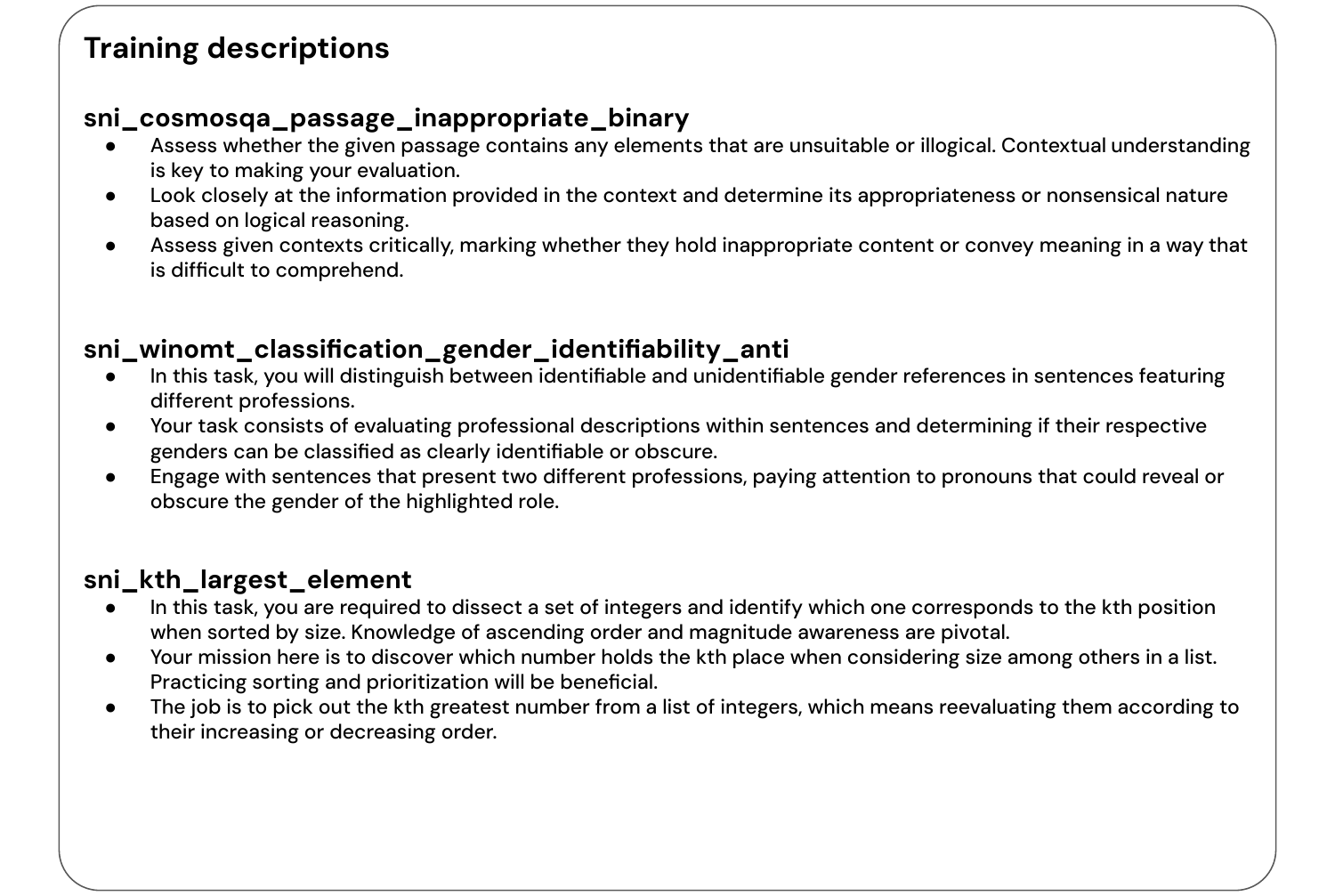}
    \caption{Examples of training descriptions from three SNI training tasks.}
    \label{fig:example_training_descriptions}
\end{figure}
\begin{figure}[H]
    \centering
    \includegraphics[page=4,trim={0cm 1cm 0cm 1cm},clip,width=1.0\linewidth]{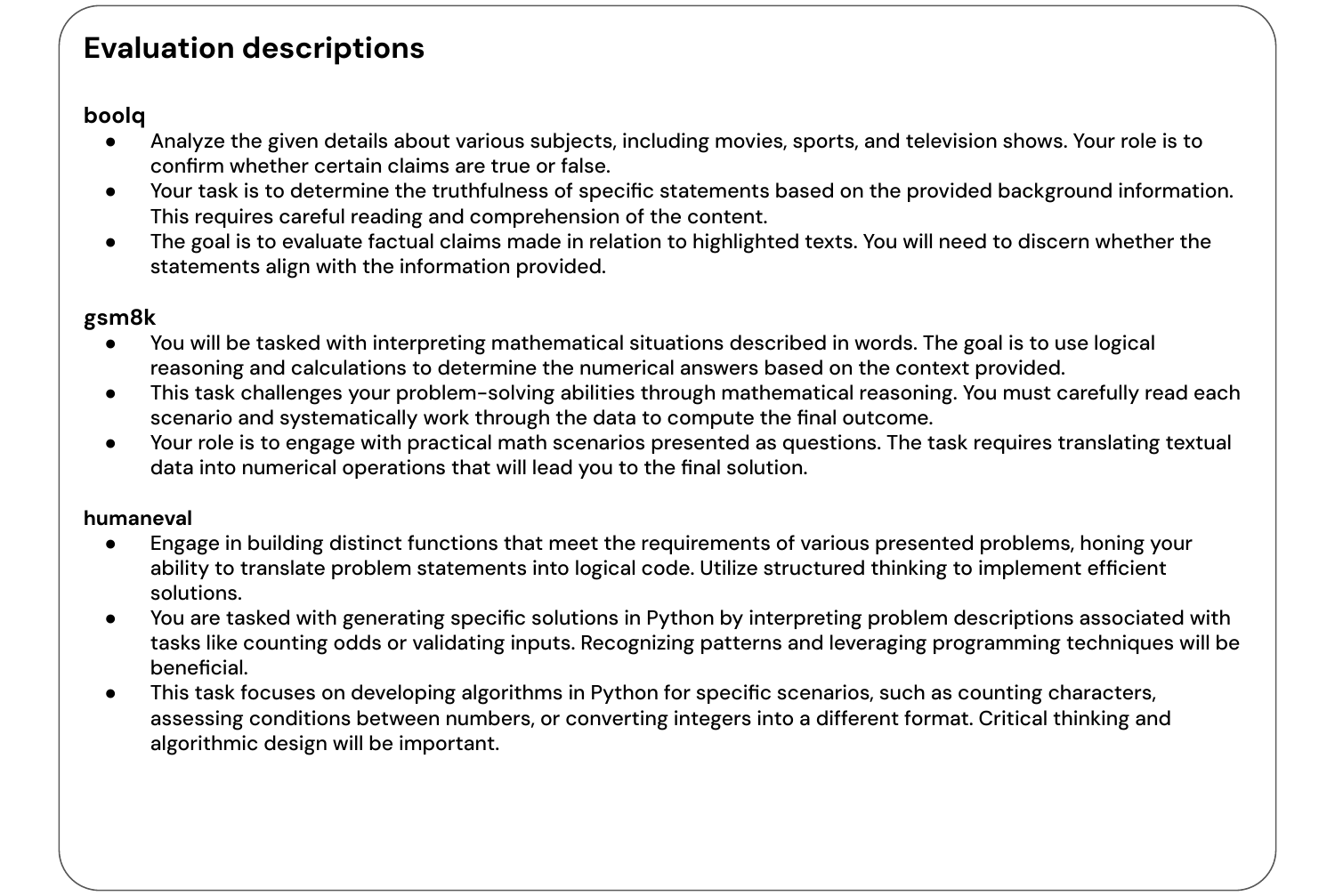}
    \caption{Task descriptions of the benchmark tasks: boolq, gsm8k, and humaneval.}\label{fig:eval_descs1}
\end{figure}
\begin{figure}[H]
    \centering
    \includegraphics[page=5,trim={0cm 1cm 0cm 1cm},clip,width=1.0\linewidth]{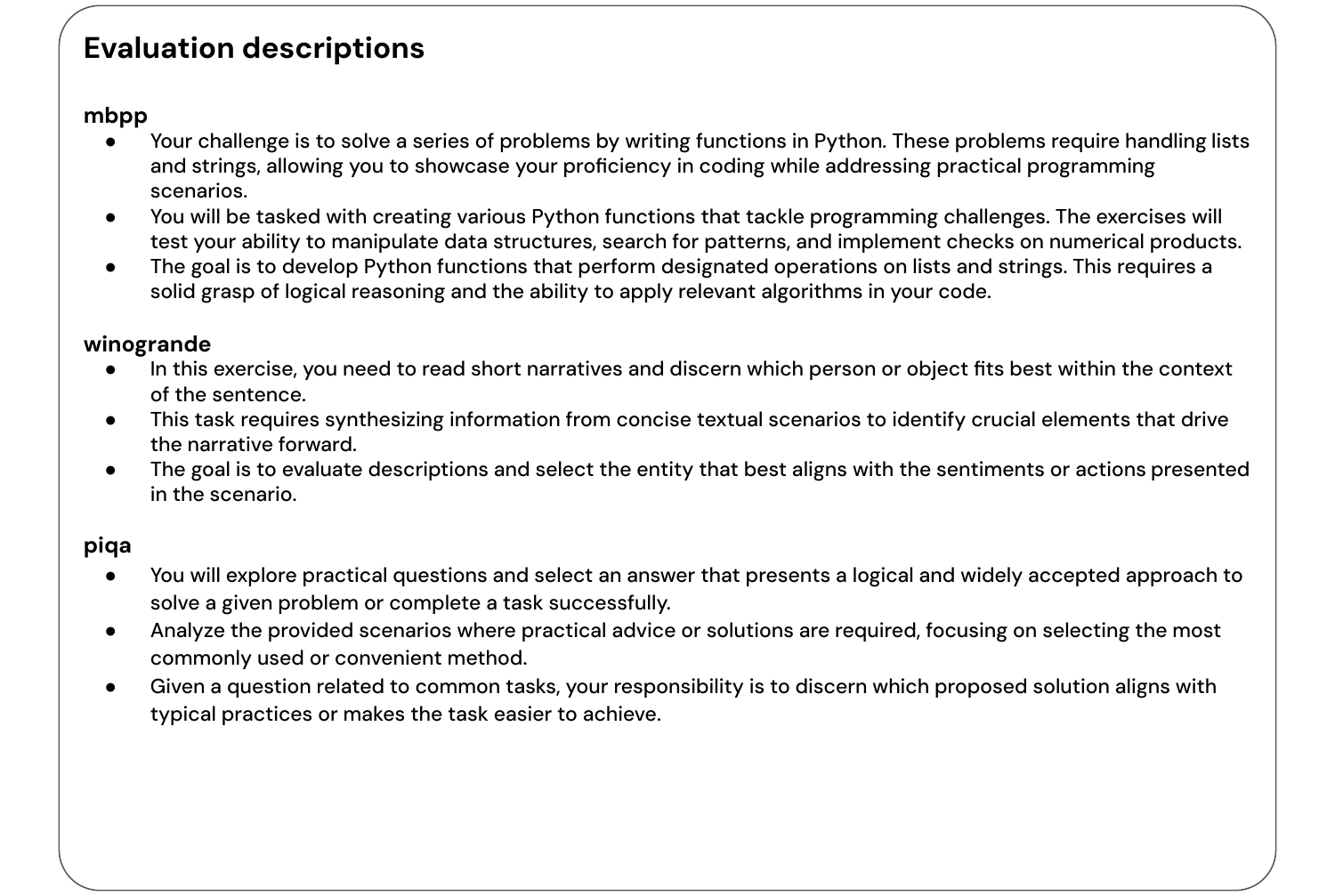}
    \caption{Task descriptions of the benchmark tasks: mbpp, winogrande, piqa}\label{fig:eval_descs2}
\end{figure}
\begin{figure}[H]
    \centering
    \includegraphics[page=6,trim={0cm 1cm 0cm 1cm},clip,width=1.0\linewidth]{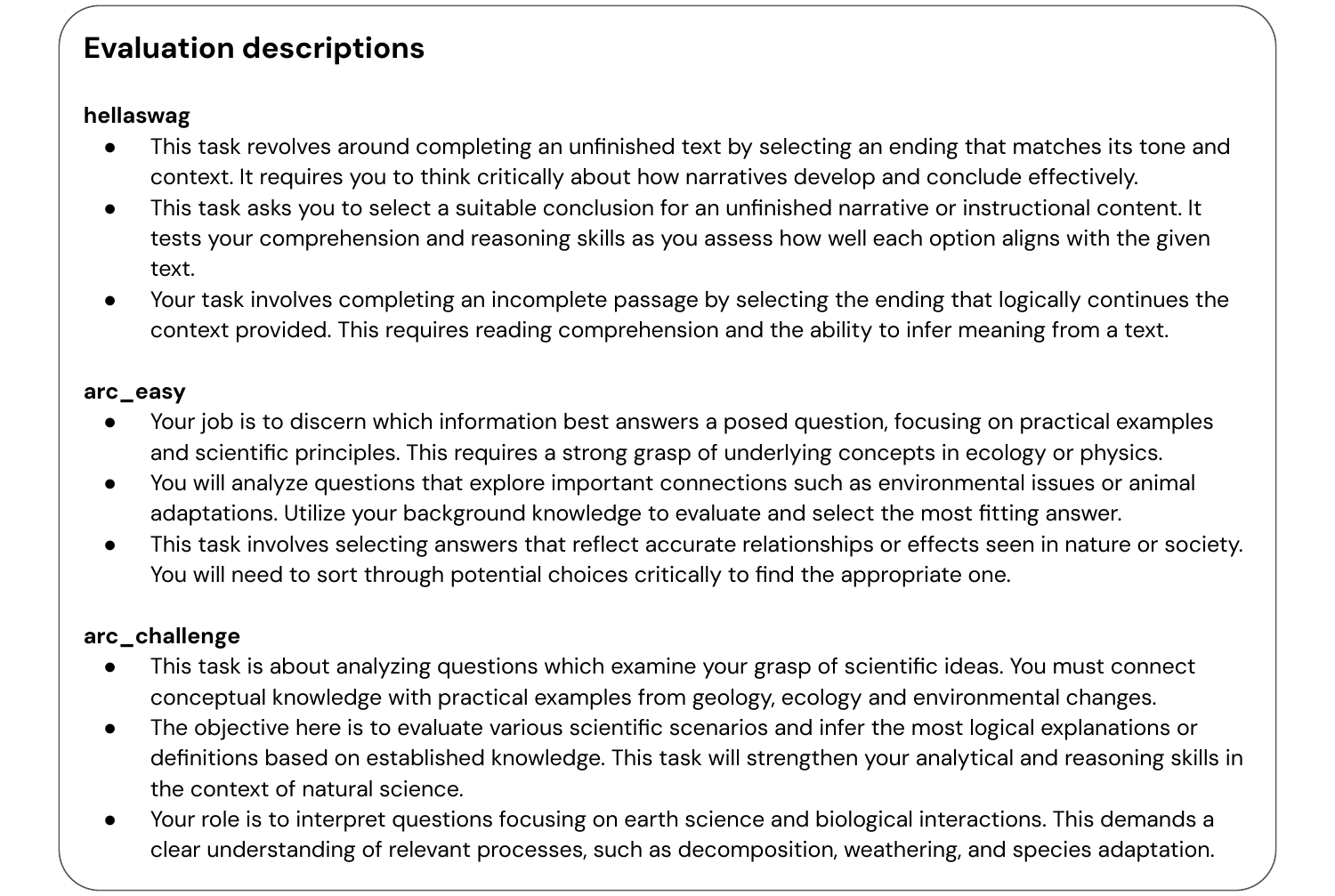}
    \caption{Task descriptions of the benchmark tasks: hellaswag, arc\_easy, arc\_challenge}\label{fig:eval_descs3}
\end{figure}
\begin{figure}[H]
    \centering
    \includegraphics[page=7,trim={0cm 1cm 0cm 1cm},clip,width=1.0\linewidth]{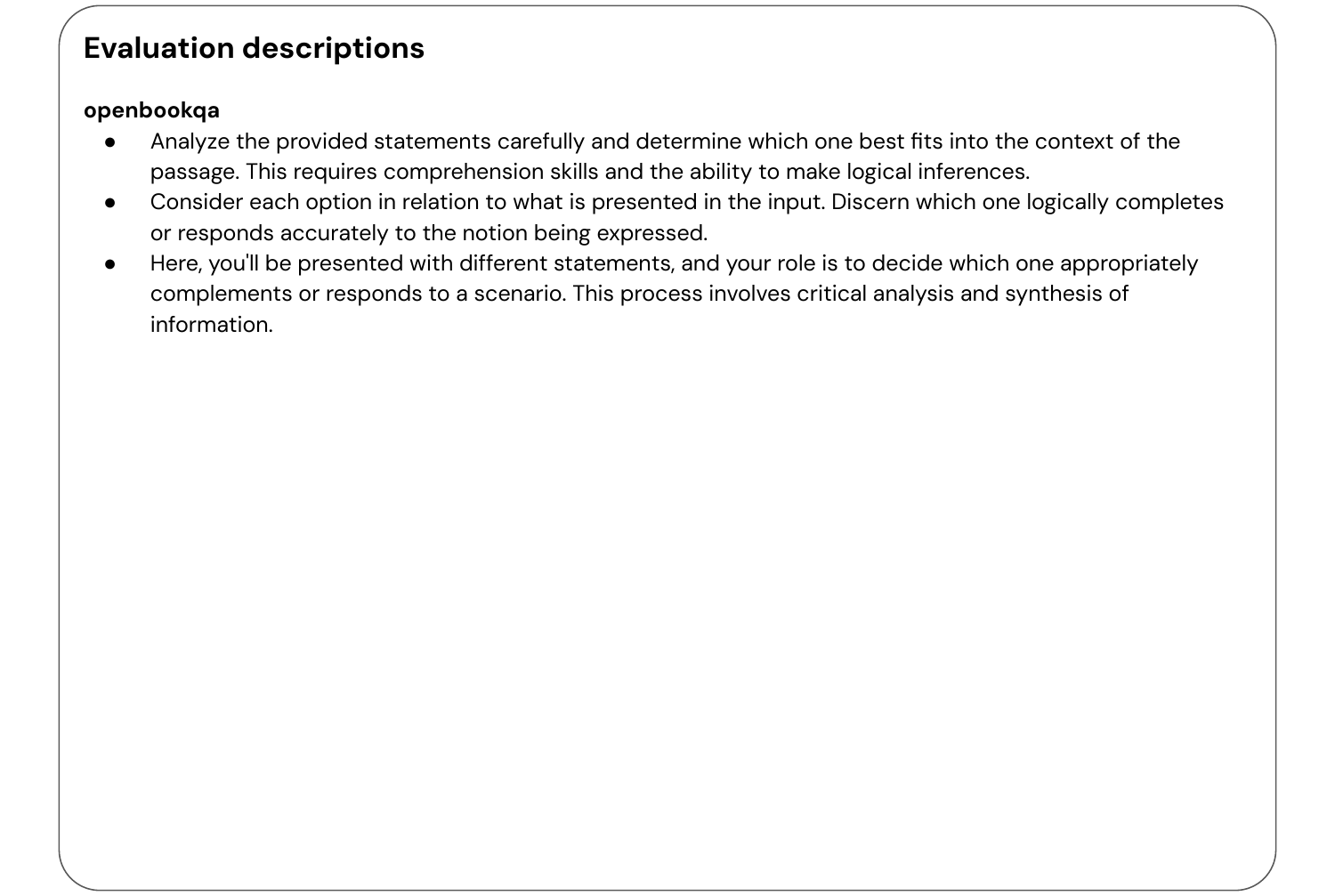}
    \caption{Task descriptions of the benchmark tasks: openbookqa}\label{fig:eval_descs4}
\end{figure}
\begin{figure}[H]
    \centering
    \includegraphics[page=8,trim={0cm 13cm 0cm 0cm},clip,width=1.0\linewidth]{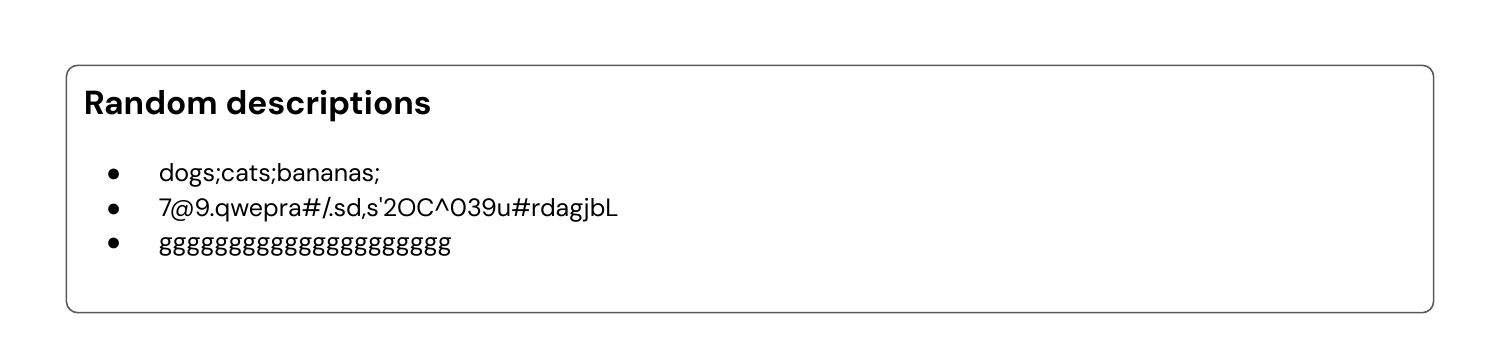}
    \caption{Random descriptions}\label{fig:random_descriptions}
\end{figure}

\section{Scaling Number of Descriptions per Task}\label{app:scaling_num_descs}
\begin{figure}[H]
    \centering
    \includegraphics[width=0.9\linewidth]{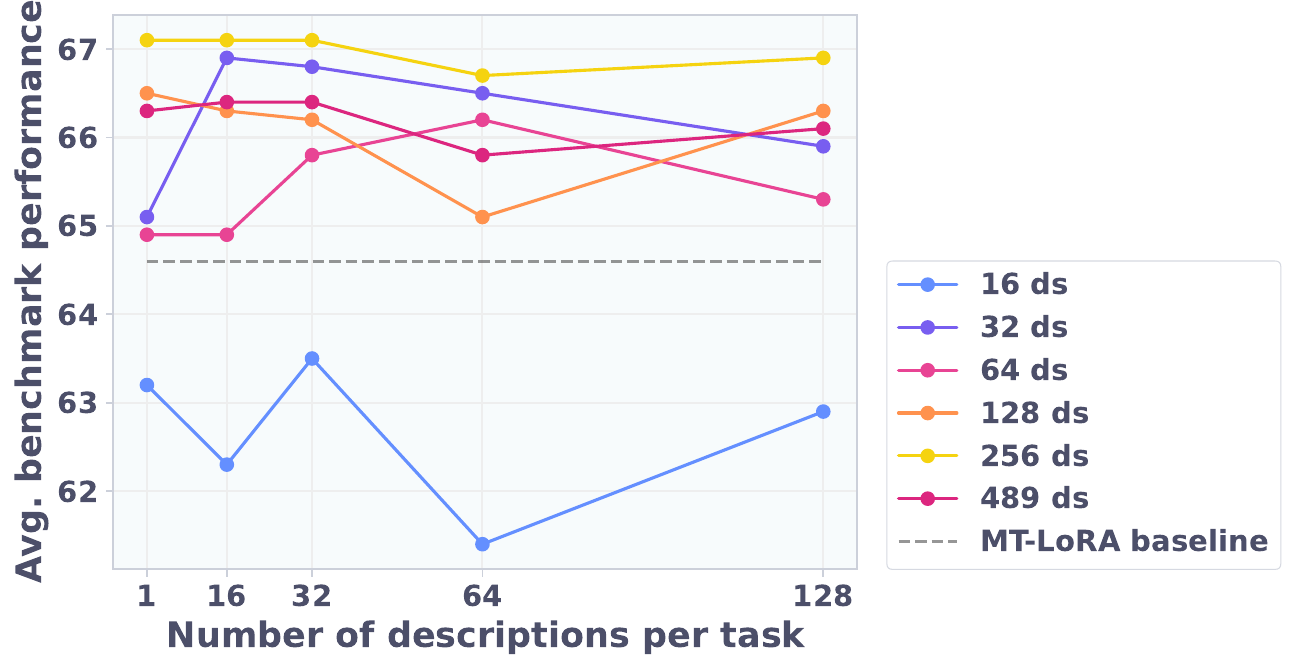}
    \caption{Zero-shot benchmark performance of SFT-trained \ours{} with varying number of descriptions per training task.}
    \label{fig:vary_num_descs}
\end{figure}
\cref{fig:vary_num_descs} shows mixed results on the benchmark performance when varying the number of descriptions per training task. For consistency, we always train \ours{} with 128 descriptions per training task.

\end{document}